\documentclass{ieeeaccess}
\usepackage{amsmath,amssymb,amsfonts}
\usepackage{algorithmic}
\usepackage{graphicx}
\usepackage{textcomp}
\usepackage{arydshln}

\usepackage{hyperref}

\usepackage{booktabs}
\ifCLASSOPTIONcompsoc\usepackage[caption=false, font=normalsize, labelfont=sf, textfont=sf]{subfig}
\else
\usepackage[caption=false, font=footnotesize]{subfig}
\usepackage[%
backend=biber,
url=true,
bibencoding=utf8,
bibstyle=ieee,
]{biblatex}
\AtEveryBibitem{
  \clearfield{doi}
  \clearfield{isbn}
  \clearfield{issn}
  \clearfield{month}
  \ifentrytype{inproceedings}{
    \clearfield{arxivId}
    \clearfield{archivePrefix}
    \clearfield{eprint}
    \clearfield{url}
  }{}
  \ifentrytype{article}{
    \clearfield{arxivId}
    \clearfield{archivePrefix}
    \clearfield{eprint}
    \clearfield{url}
  }{}
  \ifentrytype{report}{
  }{}
}
\renewbibmacro{in:}{%
  \ifentrytype{inproceedings}{%
    \setunit{}
    \addperiod\addspace In \textit{Proc.\ of the}}%
  {\printtext{\bibstring{in}\intitlepunct}}
}
\DeclareSourcemap{
  \maps{
    \map{ 
      \step[fieldsource=url,
      match=\regexp{\{\\\_\}|\{\_\}|\\\_},
      replace=\regexp{\_}]
    }
    \map{ 
      \step[fieldsource=url,
      match=\regexp{\{\$\\sim\$\}|\{\~\}|\$\\sim\$},
      replace=\regexp{\~}]
    }
    \map{ 
      \step[fieldsource=url,
      match=\regexp{\{\\\x{26}\}},
      replace=\regexp{\x{26}}]
    }
  }
}

\def\BibTeX{{\rm B\kern-.05em{\sc i\kern-.025em b}\kern-.08em
    T\kern-.1667em\lower.7ex\hbox{E}\kern-.125emX}}

\begin{document}
\history{Date of publication xxxx 00, 0000, date of current version xxxx 00, 0000.}
\doi{}

\title{Image-based scene recognition for robot navigation considering traversable plants and its manual annotation-free training}
\author{\uppercase{Shigemichi Matsuzaki}\authorrefmark{1}, \IEEEmembership{Graduate Student Member, IEEE},
  \uppercase{Hiroaki Masuzawa\authorrefmark{1}},\\
  and \uppercase{Jun Miura\authorrefmark{1}}, \IEEEmembership{Member, IEEE}.
}
\address[1]{Department of Computer Science and Engineering,
  Toyohashi University of Technology, Toyohashi 441-8580, Japan }
\tfootnote{
  This work is supported by Knowledge Hub Aichi Priority
  Research Project (third term). The work of Shigemichi Matsuzaki is also supported
  in part by the Leading Graduate School Program, “Innovative
  program for training brain-science-information-architects by
  analysis of massive quantities of highly technical information
  about the brain,” by the Ministry of Education, Culture,
  Sports, Science and Technology, Japan.    
}

\markboth
{Matsuzaki \headeretal: Scene recognition for robot navigation considering traversable plants and manual annotation-free training}
{Matsuzaki \headeretal: Scene recognition for robot navigation considering traversable plants and manual annotation-free training}

\corresp{Corresponding author: Shigemichi Matsuzaki (e-mail: matsuzaki@aisl.cs.tut.ac.jp).}

\date{\today}

\begin{abstract}
  This paper describes a method of estimating the traversability of
  plant parts covering a path and navigating through them
  for mobile robots operating in plant-rich environments.
  Conventional mobile robots rely on scene recognition methods that consider
  only the geometric information of the environment. 
  Those methods, therefore, cannot recognize paths as traversable
  when they are covered by flexible plants. 
  In this paper, we present a novel framework of image-based scene recognition
  to realize navigation in such plant-rich environments.
  Our recognition model exploits a semantic segmentation branch for
  general object classification
  and a traversability estimation branch for estimating pixel-wise traversability.
  The semantic segmentation branch is trained using
  an unsupervised domain adaptation method and
  the traversability estimation branch is trained with
  label images generated from the robot's traversal experience
  during the data acquisition phase, coined {\it traversability masks}.
  The training procedure of the entire model is, therefore, free from manual annotation.
  In our experiment, we show that the proposed recognition framework 
  is capable of distinguishing traversable plants
  more accurately than a conventional semantic segmentation with
  traversable plant and non-traversable plant classes, and an existing 
  image-based traversability estimation method.
  We also conducted a real-world experiment and confirmed that 
  the robot with the proposed recognition method successfully navigated
  in plant-rich environments. 
\end{abstract}


\titlepgskip=-15pt

\maketitle

\begin{figure*}[tb]
  \subfloat[Network architecture]{
  \includegraphics[width=0.50\hsize]{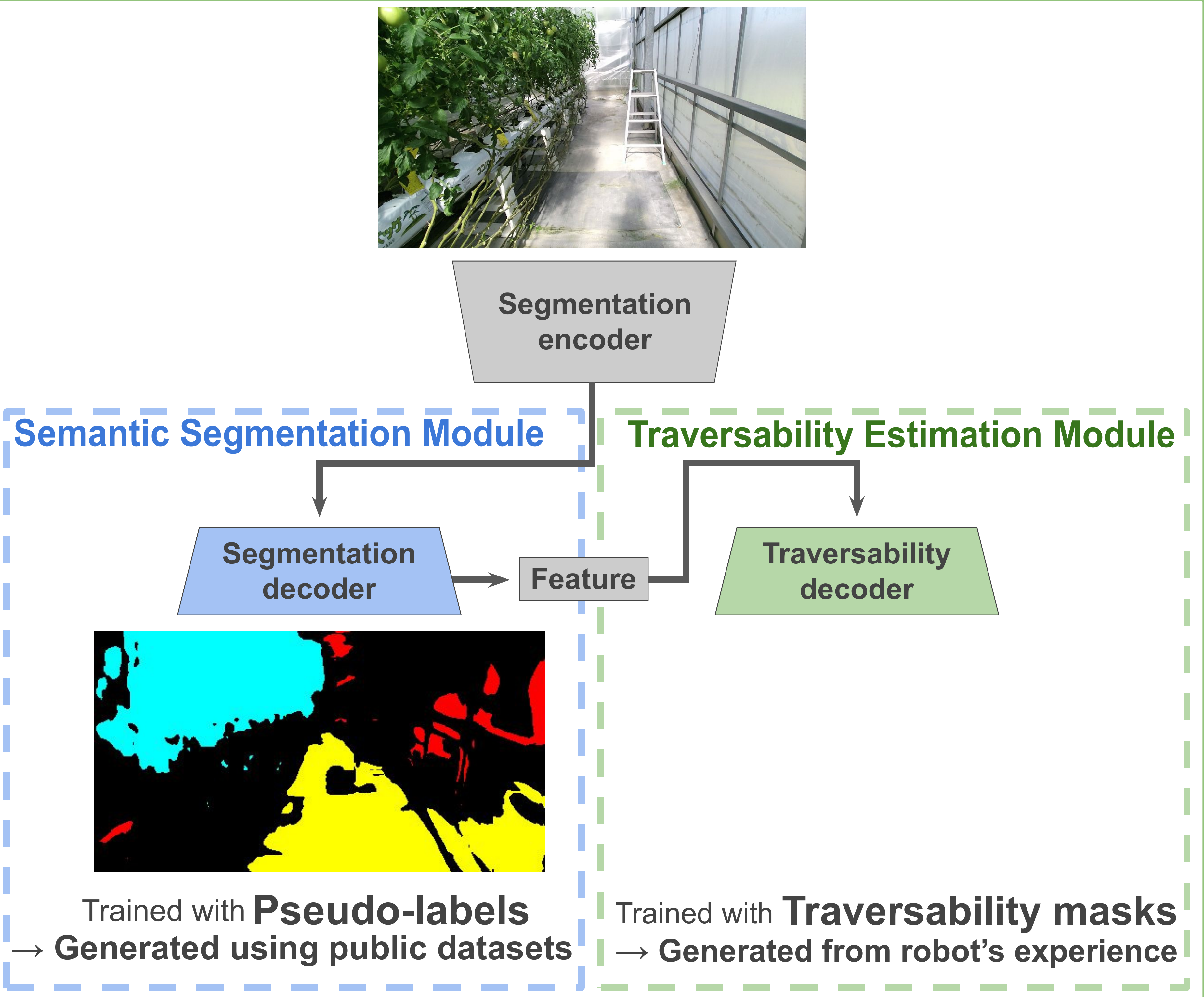}
  \label{fig:overview_training_process}}
  \subfloat[Navigation pipeline]{
  \includegraphics[width=0.50\hsize]{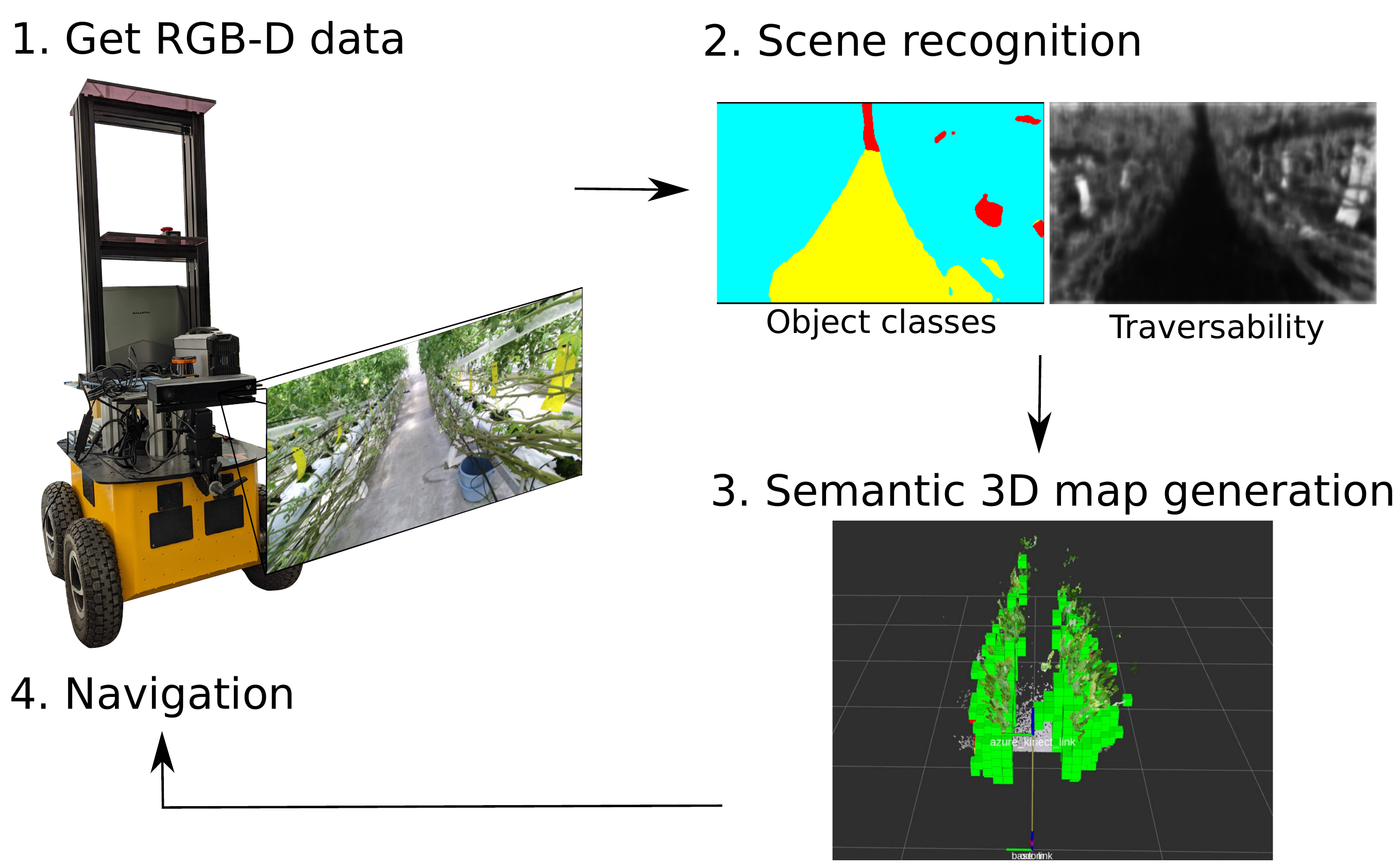}
  \label{fig:overview_navigation_pipeline}}
  \caption{Overview of the proposed method. (a) Our image-based recognition module
  consists of two modules: Semantic Segmentation Module (SSM) for pixel-wise
  general object classification, and Traversability Estimation Module (TEM) for
  pixel-wise traversability estimation. SSM is trained by a pseudo-label learning method.
  TEM is trained with the {\it traversability masks} generated based on the robot's experience.
  (b) During the robot navigation, object classes and the traversability are predicted,
  fused with depth data and projected in the 3D space to build a semantic 3D map.
  The voxels predicted as plants and with high traversability are not considered as obstacles
  and thus the robot is able to traverse paths covered by traversable plant parts.}
  \label{fig:overall_process}
\end{figure*}

\section{INTRODUCTION}

The mobile robot technologies have come nearly to the level of practical realization
and commercialization in some situations.
For example, self-driving cars are already being tested in public areas \cite{Waymo2020,TierIV2020}.
Another example 
is the service robots
operating in public places such as airports and stores \cite{Pandey2018,Triebel2016}.
Those applications mainly target structured environments
such as urban areas and the inside of buildings.
In plant-rich environments such as agricultural fields and forest paths, however,
it is more difficult for robots
to autonomously navigate than in structured environments
because the paths are possibly covered by plant parts such as branches and leaves,
which, though, can be driven through by the robots.
Most of conventional mobile robots only consider the presence of objects and
do not consider such {\it traversable objects}.

One of such environments is greenhouses and agricultural fields.
Traditionally, in such environments, plants are extracted from sensor readings
and then two plant rows that form a path are detected
via simple line fitting algorithms \cite{Bakker2008,Xue2012,Malavazi2018a,Winterhalter2020,Le2020}.
Those methods assume that paths are straight and open, and not covered by any plant parts.
A recent work \cite{Kahn2021} applies an end-to-end policy learning
to learn a function that directly maps an image to an 
appropriate control.
Although their method succeeded to navigate in outdoor scenes with tall grass,
it requires ``trial and error'' experiences achieved via 
a robot's random walk in the real environments, 
which are difficult to acquire in environments like greenhouses
and other environments with objects that the robot must not collide with.
Imitation learning (IL) is another possible approach for end-to-end policy learning
that takes a human's demonstration of control for supervision.
However, IL requires abundant amount of high-quality demonstrations given by experts,
which are not easy to collect.
Moreover, even when using learned control policies,
explicit scene recognition is still important for
safe deployment of navigation systems in such environments.
Therefore, we aim at an image-based scene recognition 
considering traversable plants for robot navigation
in plant-rich environments such as greenhouses.

In this paper, we present a novel framework of scene recognition and
robot navigation that takes into account traversable plants covering the paths.
To recognize objects as traversable, 
it is not sufficient to just recognize plants 
because plants have both traversable parts
such as branches and leaves, and non-traversable parts such as stems.
Datasets with such fine-grained classes are not publicly available
to the best of our knowledge.
To provide information of \textit{traversability} of objects,
which cannot be acquired from existing datasets, 
we use label data indicating the regions with objects 
that the robot has traversed 
during data acquisition phases, coined {\it traversability masks}.
Since the traversability masks are generated based on 
a limited number of the robot's traversal experiences,
all traversable regions are not labeled but some of them are left unlabeled
(See Fig. \ref{fig:trav_mask_example}).
Directly using such masks as labels of traversable plants
will confuse the training.
To take full advantage of the incomplete information of the traversability masks,
we present a deep neural network (DNN) architecture and its training method
inspired by PU (positive and unlabeled) learning \cite{Elkan2008},
where a model is trained with labeled positive examples and 
unlabeled examples including both positive and negative ones.

The overview of the proposed method is shown in Fig. \ref{fig:overall_process}.
The proposed DNN architecture
consists of Semantic Segmentation Module (SSM)
for pixel-wise general object classification and
Traversability Estimation Module (TEM) for estimating traversability
which indicates how likely that
each pixel can be traversed by robots.
SSM is trained without a manually labeled greenhouse dataset
by means of an unsupervised domain adaptation (UDA) method
proposed by the authors \cite{Matsuzaki2020}.
TEM is then trained using the features from SSM as inputs
and the traversability masks.
For the training of TEM, 
we introduce a training method inspired by PU learning \cite{Elkan2008}
in order to take full advantage of the information
given by the traversability masks.
The overall network is therefore trained without manual annotation of the training images.

The prediction results (the object class and the traversability of each pixel)
are then projected to the 3D space to build a semantic 3D voxel map
used in navigation.
For each data frame, the prediction results (observations)
are temporally fused with the prior probabilities
of the object classes and the traversability
in each voxel using Bayes' update.
By treating as free spaces voxels where the probability
of plant is the highest and the traversability 
is greater than a threshold, 
the robot is able to navigate through traversable plant parts 
covering the paths.

The contributions of this paper are as follows:
\begin{enumerate}
\item A novel framework of scene recognition considering
both general object classes and the traversability of objects
for robot navigation in plant-rich environments.
\item A manual annotation-free training method for the scene recognition model.
\item A PU learning-based training for traversability estimation which 
requires only positive labels given to a part of traversable regions in images.
\item Applying the scene recognition model to the navigation tasks
in a real greenhouse with plants partially covering the paths.
\end{enumerate}

\section{RELATED WORK}

\subsection{Traversability estimation}

\subsubsection{Geometry-based methods}

Traverasability estimation is a crucial task in robot navigation.
Traditionally, rigid obstacles and negative obstacles such as ditches and holes
are detected using geometric information from range sensors such as 
laser range finders (LRFs), stereo cameras, and RGB-D sensors.
In unstructured scenes, many studies have focused on estimating
the traversability of terrains
by analyzing the steepness, roughness and so on
\cite{Bogoslavskyi2013,Zhu2013a,Martinez2020a,Zhou2021}.
Bogoslavsky et al. \cite{Bogoslavskyi2013} 
estimated the traversability of indoor environments using Kinect sensor
by analyzing the normals. 
Zhu et al. \cite{Zhu2013a} exploited disparity information from a stereo camera 
to estimate traversable regions consisting of a flat plane.
Mart\'{i}nez et al. \cite{Martinez2020a} used static obstacle detection 
and terrain analysis using 2D LRF
in their reactive navigation scheme.
The near-to-far approach is also a popular approach, where 
the terrain traversability is estimated
within the measurement range of a range sensor
and the estimated traversability is then propagated 
to farther areas based on the similarity of image features
\cite{Happold2006,Hadsell2008,Wang2010}.

The researches mentioned above are for estimating the traversability of 
the terrain, i.e., whether the robot can drive over the terrain.
Unlike those methods, we deal with the estimation of
the traversability of objects such as plant parts 
hanging over and covering the paths,
i.e., whether the robot is allowed to touch and push them while navigating.

\subsubsection{Semantics-based methods}

Other methods exploit object semantics estimated from RGB images
\cite{Zhou2012,Matsuzaki2018b,Maturana2018,Gao2021,Guan2021a,Hosseinpoor2021}.
The rich information of RGB images
such as colors and texture allows for estimating 
object semantics such as object class and traversability from the appearance.
Zhou et al. \cite{Zhou2012} trained a terrain classifier on visual features 
with class labels automatically given using structural information from a 3D LiDAR. 
Matsuzaki et al. \cite{Matsuzaki2018b} used examples of 
traversable path lines given by human annotators to 
train a probabilistic model to classify the image regions 
into either of the object classes: building, road, or grass.
In the last decade, DNNs played a crucial role 
to provide high recognition accuracy and generalization ability.
Maturana et al. \cite{Maturana2018} generated a 2.5D semantic map using 
DNN-based semantic segmentation for path planning of autonomous off-road vehicles.
Gao et al. \cite{Gao2021} proposed a fine-grained traversability estimation 
to deal with different level of traversability of the terrain.
Guan et al. \cite{Guan2021a} proposed a hybrid method of DNN-based semantic segmentation
and analysis of geometric information for autonomous navigation of excavators.

None of the aforementioned methods, however, addressed
traversable objects such as plants covering the paths,
which are the main target of our method.
Although the authors' previous work \cite{Matsuzaki2018}
constructed a semantic 3D map inside a greenhouse,
traversable and non-traversable plants 
are not distinguished and robot navigation was not realized.
In addition, those DNN-based methods to estimate 
object semantics require an abundant amount of 
manual annotation by humans,
while our method provides manual annotation-free training.

\subsubsection{Self-training based on experience}

Using information of robot's traversal is a preferred approach for 
identifying traversable regions in recent studies.
Barnes et al. \cite{Barnes2017} used trajectories of an autonomous vehicle for
automatically labeling images for semantic segmentation.
Similarly, Wellhausen et al. \cite{Wellhausen2019} proposed to use footprints
of a legged robot to label images and showed that sufficient
performance can be achieved even with such sparse labels.
Lee et al., \cite{Lee2021} trained a multi-layer perceptron (MLP)
on geometric features in a self-training manner 
with data labeling using both a robot's trajectory
and structural information from an elevation map.
Those methods are, however, not applicable to traversable objects hanging over
three-dimensionally.
Some studies consider the traversability of objects 
as an affordance depending on 
both the environment and the robots. 
Kim et al. \cite{Kim2006} trained a classifier with simple hand-crafted features
of image patches by exploiting a robot's experience of 
successful and unsuccessful traversals
acquired by its interaction with objects.
In another work \cite{Kim2007}, they proposed
a superpixel-based image classification 
to improve the accuracy of region segmentation.
The studies by Kim et al. share the most similar challenge with our work
in terms of considering the traversability of plants.

%
%
Recently, methods based on Reinforcement Learning (RL) and Imitation Learning (IL)
have also been presented.
Kahn et al. \cite{Kahn2021} applied in outdoor robot navigation 
such an end-to-end policy learning based on
robot's experience of trial and error gathered through its random walk motion
in real environments.
Despite their success, we claim that 
such a training method is not suitable
for navigation in greenhouses or other plant-rich environments
because there may be objects that the robot must not hit such as 
crop rows and equipments, or steep slopes beside the path.
Random motions in such environments may lead to a fatal failure.
IL methods instead exploit high-quality demonstrations by human experts.
Such demonstrations are, however, not always available.
Even when a kind of policy learning can be used,
explicit scene recognition is still necessary for ensuring the safety.
In this sense, our proposed method can be rather complementary to those methods.
%
%

\subsection{Agricultural mobile robots}

Agricultural mobile robots are one of the practical and actively studied
applications of mobile robots in plant-rich environments.
Many of currently used agricultural robots move on rails or heating pipes
equipped on the paths \cite{Grimstad2018,SweeperWeb2018}.
Robotic systems with such locomotion is able to move stably,
while movable areas for operation are limited.
In addition, such systems require installation of the rails when not installed.
More flexible robot operation is beneficial for extending the field of tasks that 
the robots can automate.

In outdoor, precise localization by GNSS is adopted in various studies \cite{Nagasaka2004,Cantelli2019}.
In indoor fields such as greenhouses, however, such a method is not available
due to noise imposed by surrounding obstacles.
Therefore, a popular approach 
is recognizing a traversable path using sensors equipped on the robot.
The sensors used in path recognition include ultrasonic sensors \cite{Mandow1996,Dong2011},
LIDARs \cite{Malavazi2018a,Harik2018,Le2020} and
monocular / stereo cameras \cite{Younse2007,Bakker2008,Xue2012,Winterhalter2020}.

Most of those methods detect two plant rows that form a path
with a combination of plant detection based on sensor readings,
and line fitting algorithms such as
Hough transformation \cite{Bakker2008,Winterhalter2020}
and iterative methods like RANSAC \cite{Le2020,Malavazi2018a}.
Those methods assume that the lines of plants are visible.
However, this is not the case in environments where some plant parts
cover the path.
In one of pioneering work by Mandow et al. \cite{Mandow1996}, it has been reported that
while they succeeded to autonomously operate a mobile robot using an ultrasonic sensor,
some failure cases observed where the robot recognized plant parts overhanging the path
as obstacles and stopped the navigation.
To the best of our knowledge, the problem of recognizing traversable plant parts
covering the paths has not been explicitly addressed.

\subsection{Semantic segmentation}

Semantic segmentation is a task to assign an object class to each pixel of an image.
Long et al. \cite{Long2015} first proposed to convert classification networks
into fully convolutional network to produce a probability map for each pixel.
Based on ResNet proposed by He et al. \cite{He2016} as a backbone network,
very deep networks have been shown to achieve high accuracy \cite{Zhao2017a,Chen2018-deeplab}.
While it is proven to be effective to stack layers in depth to get a high accuracy,
those models impose high computational cost.
For real-time visual tasks, computationally efficient architectures have also been studied,
targeting the applications such as robotics and autonomous driving systems
\cite{Badrinarayanan2017,Ronneberger2015,Paszke2016}.
Recent approaches further improved the accuracy while maintaining the efficiency by introducing 
more efficient convolution approaches such as factorized convolution \cite{Romera2018},
depth-wise separable convolution \cite{Sandler2019},
efficient spatial pyramid \cite{Mehta2018,Mehta2019} etc.

In our previous work, we have shown that a semantic segmentation network for
greenhouses can be trained without requiring hand-labeled images,
by utilizing multiple publicly available datasets of urban scenes and
employing an unsupervised domain adaptation (UDA) method \cite{Matsuzaki2020}.

\subsection{PU learning}

PU (positive and unlabeled) learning is a problem where
part of the data is labeled as positive, and the rest is unlabeled and
could be either positive or negative.
Unlike the setting where positive and negative data are assumed,
PU setting cannot be solved by just applying a simple classification
of positive and unlabeled data since the unlabeled data include
positive ones.
Elkan and Noto \cite{Elkan2008} showed that
it is possible to estimate the label probabilities by 
modeling the probability that given samples are labeled.
As a practical application, 
Yang et al. \cite{Yang2020} applied PU learning framework in 
object detection task where some of objects are not labeled.

In our work, we exploit binary-labeled images where pixels that the robot has
traversed are labeled, and the rest is left unlabeled.
We can treat the labeled pixels as ``traversable'' based on the robot's traversals,
whereas it is not always true that the unlabeled pixels are ``non-traversable'',
since there is a substantial amount of unlabeled pixels of traversable objects
as later shown in \ref{sec:traversability_mask}.
This problem is exactly an example of PU learning.
We therefore introduce a training method
inspired by Elkan and Noto \cite{Elkan2008}
to the training of TEM in \ref{sec:network_architecture_tem}.

\section{PROPOSED METHOD}

Fig. \ref{fig:overall_process} 
shows the overview of the proposed framework.
We employ two network modules for the scene recognition:
Semantic Segmentation Module (SSM)
for semantic segmentation, and
Traversability Estimation Module (TEM) for traversability estimation.
SSM is trained by an unsupervised domain adaptation method.
For the training of TEM, we generate traversability masks
based on the robot's traversal experience.
The network is then trained with the generated traversability masks
(Fig. \ref{fig:overview_training_process}).
Predictions of the trained network are projected
to the 3D space to build semantic local 3D map
that indicates the object classes and traversability of the regions around the robot
and the navigation is operated based on the map (Fig. \ref{fig:overview_navigation_pipeline}).

\subsection{Traversability mask}
\label{sec:traversability_mask}

The traversability masks are label data that indicate the image regions traversed
by the robot or a human during the data acquisition phase.
Using the traversability masks, we automatically label the regions of images dominated by
traversable plant parts to train a network so that it can distinguish
traversable and non-traversable plants based on the robot's experience.

We first acquire RGB-D data by manually operating a robot or 
walking inside the target greenhouse with an RGB-D sensor 
moving through the plants.
We then build a 3D voxel map using RTAB-Map \cite{Labbe2018} from the RGB-D data.
After that, by utilizing the robot's path estimated in the mapping process, the voxels traversed by
the robot or the human are labeled.
Here, we approximate the shape of the robot or the human with a rectangular,
and from each observation point of the 3D voxel map,
voxels within the rectangular are labeled as traversed.
Finally, traversability masks are generated by projecting the traversability labels of the
voxels on an image plane from each sensor pose on the map.

\Figure[t!](topskip=0pt, botskip=0pt, midskip=0pt)[width=0.9999\linewidth]{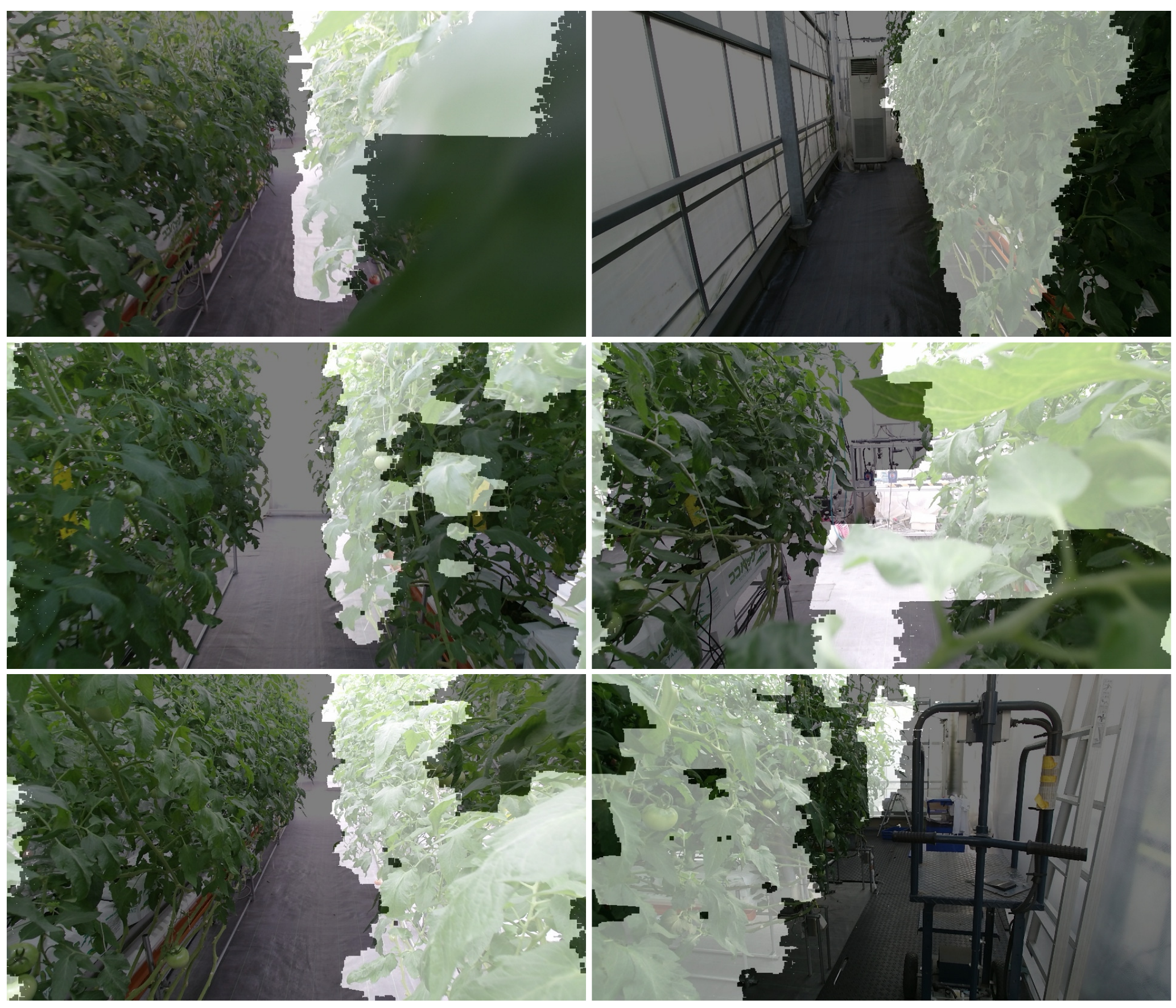}
{Examples of traversability masks. Regions traversed by the robot
are superimposed on the corresponding camera images \label{fig:trav_mask_example} }
Examples of the traversability masks are shown in Fig. \ref{fig:trav_mask_example}.
One may notice that while much of traversable plant parts such as leaves and branches 
are labeled, there is also quite a lot of unlabeled regions of traversable plants.
This is because the traversability masks are based on the limited number of 
traverses during the data acquisition phase.
This nature of the traversability masks makes it inappropriate to 
simply treat them as positive and negative examples.
By our training method inspired by PU learning \cite{Elkan2008} described in
\ref{sec:network_architecture_tem},
the traversability masks are effectively exploited
to distinguish traversable and non-traversable plants.

\subsection{Network architecture}
\label{sec:network_architecture}

In this section, we describe a network architecture 
for pixel-wise semantic segmentation and 
traversability estimation.
The overview of our network is shown in Fig. \ref{fig:overview_training_process}. 
The network consists of Semantic Segmentation Module (SSM)
for pixel-wise object label assignment and
Traversability Estimation Module (TEM) for estimating how likely that
each pixel can be traversed by robots.
SSM is first trained independently. 
TEM is then trained with intermediate features of SSM as 
inputs and traversability masks as training labels.
The reason for not training them simultaneously is to
avoid overfitting of the entire network to the traversability masks
which include unlabeled traversable regions.

Our network is based on ESPNetv2 \cite{Mehta2019} and 
we make two modifications.
First, an auxiliary segmentation branch is attached 
to the middle of the main network of ESPNetv2. 
We refer to the segmentation network including 
the auxiliary branch as SSM (Semantic Segmentation Module).
The auxiliary branch is used to estimate pixel-wise uncertainty 
for the training of SSM proposed in \cite{Matsuzaki2020}.
Second, a branch for estimating traversability, i.e., TEM is attached.
The intermediate feature from the SSM is 
fed into a $3\times3$ convolution and
the sigmoid function to produce pixel-wise probability predictions.
For the complete network architecture, see Fig. \ref{fig:network_detail}
in the Appendix.
The additional layers do not affect the computational efficiency 
and still allows for prediction at sufficient speed as explained in 
\ref{sec:navigation_in_a_greenhouse}.

The advantages of our architecture are three-fold.
Firstly, the architecture enables the PU learning-based training
to model the traversability as probability values.
Using the semantic segmentation task provides
discriminative features for the PU learning.
Secondly, by separating the tasks into semantic segmentation and traversability estimation,
the branches for each task can be trained with the maximum possible number of data.
Because the traversability masks are generated from a map, 
the number of available training data for TEM is less than 
that for SSM, which only requires RGB images.
Thanks to the network structure, 
the SSM can fully exploit available training images 
for a better performance of semantic segmentation individually from 
the training of TEM.
In fact, as shown later in \ref{sec:evaluation_of_tem},
the SSM was trained with 6689 images while the TEM was trained with 
889 images with traversability masks in our experiment.
Thirdly, the two modules provide complementary information.
Misclassifications of TEM can be refined by a prediction 
of SSM by utilizing prior knowledge that, e.g., 
artificial objects are not traversable.
We evaluate this effect in \ref{sec:quantitative_evaluation}.

\subsubsection{Semantic Segmentation Module}
\label{sec:network_architecture_ssm}

Semantic Segmentation Module (SSM) is responsible for pixel-wise 
general object classification.
Here we use three general object classes: plants,
artificial objects, and the ground.
For training SSM, we introduce a training method 
in our previous work \cite{Matsuzaki2020}.
In the method, we employ multiple publicly available datasets with
pixel-wise labels (the source datasets) and an unlabeled greenhouse dataset
(the target dataset).
First, segmentation models are trained with each source dataset.
Images from the target dataset are then fed into the models
and corresponding outputs are yielded.
If a pixel is assigned with the same object label 
from all the source models,
the label is treated as a pseudo-label of the pixel.
For training of SSM with the pseudo-labels, we adopt 
a method of adaptively weighting a loss value on each pixel
based on the uncertainty of the prediction using 
the main and auxiliary branches \cite{Zheng2021}.

\subsubsection{Traversability Estimation Module}
\label{sec:network_architecture_tem}

Traversability Estimation Module (TEM) estimates 
the probabilities that the pixels are traversable.
In the training of TEM with the traversability masks, 
we introduce the PU (positive and unlabeled) learning framework.
The purpose of PU learning is to model a probability function
$p(y=1|x)$, where $x$ is an input data and
$y\in \{0, 1\}$ is a binary label of the data $x$,
in a situation 
where a part of the positive data ($y=1$) is labeled as positive,
and the rest is unlabeled and can be either positive or negative. 
Elkan and Noto \cite{Elkan2008} state that
under the ``selected completely at random'' assumption,
meaning that the labels are given completely at random,
PU learning can be solved by modeling a classifier $g(x)$
such that $g(x)=p(s=1|x)$,
where $s$ denotes a random variable indicating that
data $x$ is labeled if $s=1$.
Here, note that $y=1$ if $s=1$ but not vice versa.
The equation can then be transformed as follows, given
the ``selected completely at random'' assumption, i.e., $p(s=1|y=1,x)=p(s=1|y=1)$:
\begin{eqnarray}
  g(x)&=&p(s=1|x) \nonumber \\
  &=&p(y=1\land s=1|x) \nonumber\\
  &=&p(y=1 | x)p(s=1|y=1,x) \nonumber \\
  &=&p(y=1 | x)p(s=1|y=1).
\end{eqnarray}
Therefore, the traversable probability of a pixel $p(y=1|x)$ can be
calculated as follows:
\begin{equation}
  p(y=1|x)=\frac{g(x)}{c},
\end{equation}
where $c=p(s=1|y=1)$ which denotes the conditional probability that a positive
data is labeled.
$c$ is approximated by feeding the training data to the estimated $g(x)$
and average the probabilities of the labeled features, i.e.,
\begin{equation}
  c = p(s=1|y=1) \approx \frac{1}{n}\sum_{x\in P}g(x),
  \label{eq:c}
\end{equation}
where $P$ denotes a set of all labeled pixels in the training dataset
and $n$ is the number of elements in $P$.

In TEM, we first model the label probability $g(x)=p(s=1|x)$
by a convolution and a sigmoid activation.
The output is then divided by $c$ calculated by eq. (\ref{eq:c})
to produce $p(y=1|x)$.
The feature map from the SSM is fed in TEM.
The size of the feature map is $(B, C, H, W)$ where $B$ denotes the batch size,
$C$ denotes the channel size, $H$ and $W$
denote the height and the width, respectively.
Here we consider a vector of size $C$ as an input $x$ for the corresponding pixel.
$3\times 3$ convolution is applied to the features,
followed by a sigmoid function to scale the output in the range of $[0,1]$.
We use $3\times 3$ convolution instead of $1\times 1$ convolution 
to take information of adjacent pixels into consideration in the estimation.

Note that the ``selected completely at random'' assumption
does not strictly hold in our task.
In practice, however, we confirm that this formulation is effective
to estimate pixel-wise traversability.

%
%
\subsection{3D semantic voxel map}
\label{sec:proposed_method_3d_semantic_voxel_map}

Using the results of the recognition explained in section
\ref{sec:network_architecture},
we build 3D semantic voxel map around the robot.
At first, an RGB image acquired from the RGB-D sensor is passed to the segmentation network.
The predictions from SSM and TEM are then mapped into the 3D space
using the corresponding depth image.
The 3D space around the robot
is divided into voxels with a side of 0.1 [m], and the predictions mapped into the 3D space 
are assigned to a corresponding voxel that they fall into.
For the object class information, a histogram of the object classes is
constructed in each voxel and the class label with highest frequency 
within the voxel is used as an observation.
For the traversability information, the traversability values of the points
within a voxel are averaged and treated as an observation.

In each time step, those observations are temporally fused by Bayesian update.
The observation of the object class $z^o_t$ at time $t$ 
is fused by the following equation:
\begin{equation}
  P(l_t|z^o_t) = \eta P(z^o_t|l_{t-1})P(l_{t-1}),
\end{equation}
where $l_t$ denotes an object label of the voxel at time $t$, 
$\eta$ is a normalization term and $P(l_{t-1})$ is the prior of label $l_{t-1}$
at time $t-1$.
$P(z^o_t|l)$ is the likelihood of label $l$ with
observation $z^o_t$. 
%
%
%
%
The likelihood is calculated using greenhouse images
not used in the training of SSM as follows.
Firstly, pseudo-labels are generated for each image.
The same images are then fed in SSM and outputs of pixel-wise class label are produced.
For each object class $l$, a histogram of predicted object classes 
over all pixels of the pseudo-labels with the label $l$.
A conditional probability $P(z_o|l)$ is calculated by normalizing the histogram
and used as a likelihood function. 
The object class with the highest probability is assigned to the voxel as its label.

Similarly, the observation of traversability is fused by the following equation:
\begin{equation}
  P(\tau_t|z^{\tau}_t) = \eta P(z^{\tau}_t|\tau_{t-1})P(\tau_{t-1}),
\end{equation}
where $\tau_t\in \{0, 1\}$ denotes an event that the voxel is traversable (1) or
not traversable (0) at time $t$.
The likelihood $P(z^\tau_t|\tau)$ is calculated 
using the training images with traversability masks as follows. 
The input images are fed in TEM and outputs of traversability values are yielded.
For each label $\tau\in \{0,1\}$ in the traversability masks,
a histogram of estimated traversability values 
is calculated over all pixels with the traversability label $\tau$.
The histograms are then normalized to calculate a conditional probability
$P(z^\tau_t|\tau)$ and it is used as a likelihood.

After the integration of observed points to the semantic voxel map,
obstacle point cloud is generated by treating the voxels
whose object class is ``plant'' and
the traversability is higher than the threshold as free spaces and
the others as obstacles.
The location of a point for a voxel is the centroid of the points 
that have been accumulated in the voxel over the frames.
By feeding the obstacle point cloud to a conventional navigation module,
the navigation in plant-rich environments is realized.
For mitigating the degradation of the processing speed as the number of voxels increases,
the voxels where no point has been observed for a certain number of consecutive frames
are removed from the map.
In the experiment below, the number of frames is set to 10.

\section{EXPERIMENTS}

\subsection{Evaluation of TEM}
\label{sec:evaluation_of_tem}


\subsubsection{Used datasets}

\begin{table}
    \centering
    \caption{Greenhouse datasets used in the training. The type of each set is 
    shown in the brackets}
    \label{table:greenhouse_datasets_overview}
    \begin{tabular}{llll} \toprule
         & Train & Test & Date \\ \midrule
        A & 6684 (unlabeled)     & 33 (true trav. labels) & May 25, 2018 \\
        B & 899 (w/ trav. masks) & -            & July 12, 2019\\ \bottomrule
    \end{tabular}
\end{table}

For the training of SSM by the UDA method,
we used three source datasets: 
CamVid \cite{Brostow2009}, Cityscapes \cite{Cordts2016}, and Freiburg Forest \cite{Valada2017}, 
for generating pseudo-labels.
As target data, we used 6684 unlabeled greenhouse images from Greenhouse A dataset,
taken in a greenhouse growing tomatoes.

For the training of TEM,
we used 899 pairs of an RGB image and
the corresponding traversability mask from Greenhouse B dataset,
taken in the same greenhouse as Greenhouse A on a different date.
The data were collected through a human operator's control 
of the robot for about 20 minutes in total.
During the data collection, the robot traversed each of the paths
with plant rows on both sides only once.
The labels are therefore imbalanced as can be seen in Fig. \ref{fig:trav_mask_example}.
Approximately 40 to 50 \% of the traversable plant regions were actually labeled.
Greenhouse A and B have a different appearance due to
a different level of growth of the plants.
For the evaluation, we manually gave true labels of traversable regions 
on 33 images from Greenhouse A.
Overview of the greenhouse datasets is shown in Table \ref{table:greenhouse_datasets_overview}.

\subsubsection{Training conditions and hyperparameters}

The DNN model is implemented with PyTorch \cite{Paszke2019}
and all of the training is performed on one NVIDIA
Quadro RTX 8000 with 48GB of memory.
In the training of SSM, 
the training epoch is set to 200 with
a fixed learning rate of $5\times 10^{-5}$, 
and the batch size of 64.
TEM is then trained with weights of SSM fixed.
The number of epoch is set to 200 and the batch size is 64.
We use cyclical learning rate scheduling \cite{Smith2017}.
The initial learning rate is $5\times 10^{-5}$ 
and linearly increases by a factor of 10 in 10 epochs and
then decreases to the original value in 20 epochs.

\subsubsection{Baselines for traversability estimation}

As baseline methods of image-based traversability estimation,
we use the following methods.

\noindent
\textbf{Semantic segmentation}
We train a semantic segmentation model with ``traversable plant'' class as well as 
the classes used in the training of SSM.
A model identical to SSM is used as a semantic segmentation model.
For training, we used the same dataset as the one used in the training of TEM
(899 RGB images with traversability masks).
Pseudo-labels are generated for each image with the three classes.
Labels of the pixels with the ``plant'' class are then changed to
``traversable plant'' if the value of
the corresponding traversability mask is 1 (traversable).
As a result, the object classes used in the training are as follows: 
\textit{traversable plant, other plant, artificial object},  and \textit{ground}.
We initialize the network via pre-training with Greenhouse A dataset.
A constant learning rate of $1\times 10^{-4}$ is used and
an epoch size is 50.

\noindent
\textbf{Kim et al.}
Out of few studies that explicitly predict the traversability 
of objects such as plants,
we use Kim et al. \cite{Kim2006,Kim2007} as another baseline.
They addressed the problem of training a classifier based on a robot's experience of 
successes/failures of traversals, which shares the research target with ours.
To evaluate the baseline method, we conducted 10-fold cross validation 
using the 33 test images with ground truth labels.
We use those data instead of the training images with 
traversability masks for two reasons:
1. the baseline method assumes complete labels of 
traversable/non-traversable while the traversability masks are not 
complete and include unlabeled traversable regions.
2. the baseline method is for incremental online learning
with several image frames instead of hundreds of training data.
In the original work \cite{Kim2007}, ten consecutive frames of images 
are used for training.

\subsubsection{Results}
\label{sec:quantitative_evaluation}

\begin{figure}[tb]
    \centering
    \includegraphics[width=\hsize]{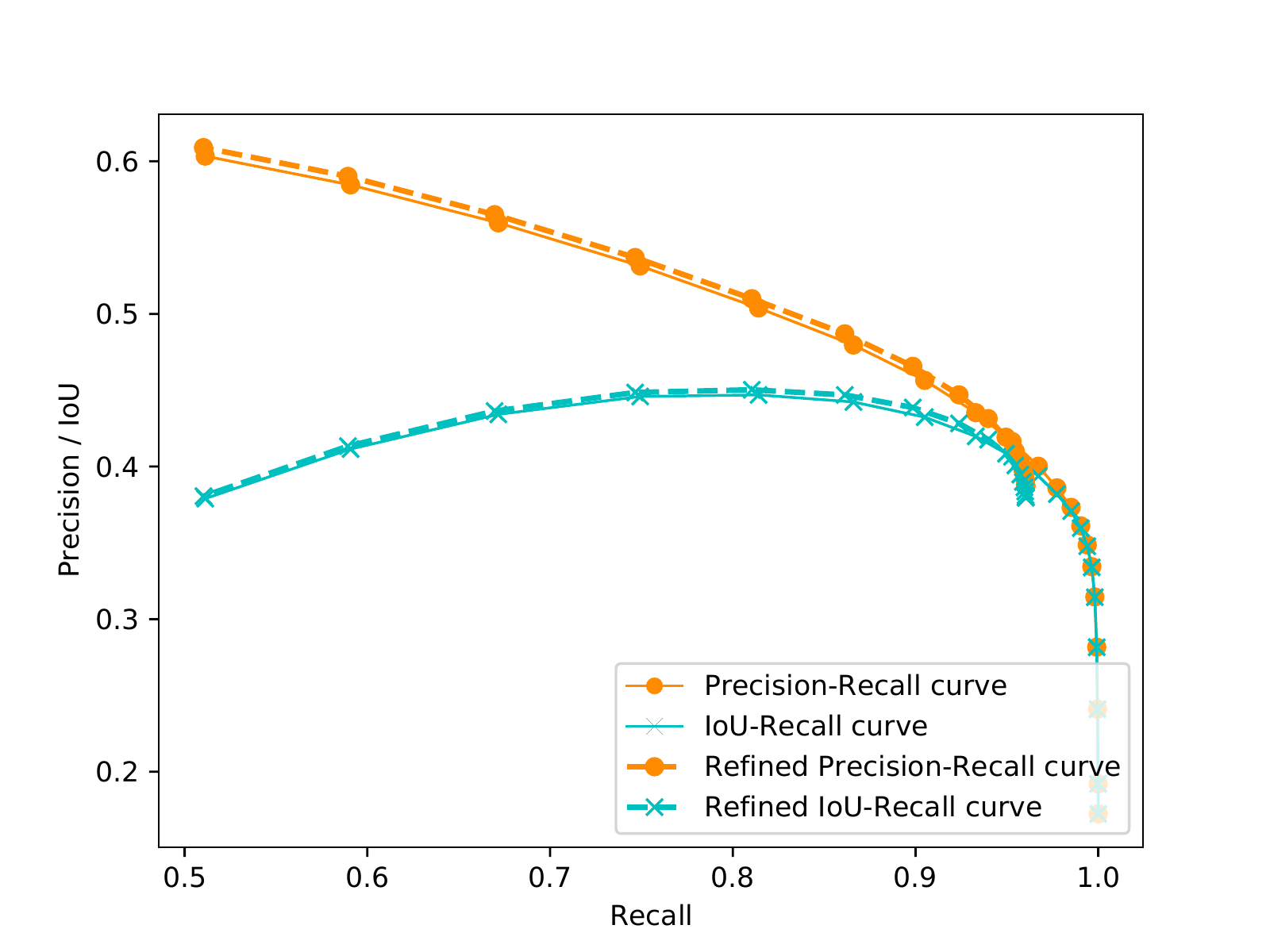}
    \caption{Precision-Recall curve and IoU-Recall curve
    of traversability estimation}
    \label{fig:pr_curve}
\end{figure}

\begin{table}
    \centering
    \caption{Performance of TEM}
    \label{table:performance_of_TEM}
    \scalebox{0.900}{ 
    \begin{tabular}{lllll} \toprule
        &IoU (\%) & Accuracy (\%) & Precision (\%) & Recall (\%) \\ \midrule
     Raw                       & 44.69       & 82.73  & 50.39 & {\bf 81.41}      \\ 
     Refined                   & {\bf 45.03} (+0.34) & {\bf 83.05} (+0.32)  & {\bf 51.00} (+0.61) & 81.04 (-0.37) \\ \hdashline 
     Segmentation              & 2.97        & 82.76  & 45.45 & 3.10     \\ 
     Kim et al. \cite{Kim2007} & 30.95       & 80.42  & 35.87 &  72.12    \\ \bottomrule
    \end{tabular}
    }
\end{table}

\begin{figure*}[tb]
    \centering
    \includegraphics[width=\hsize]{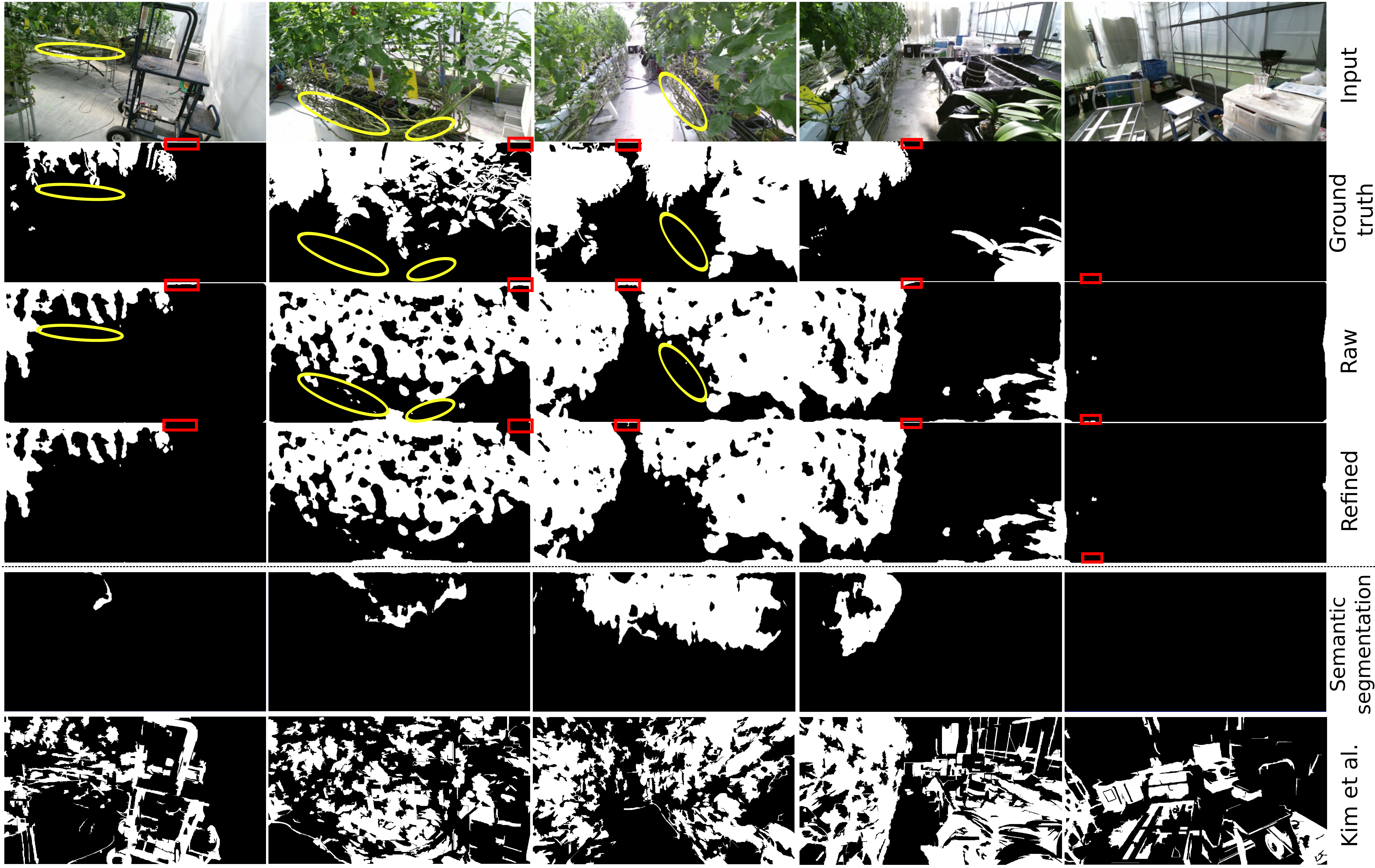}
    \caption{Prediction results of TEM. From the top: Camera images,
        Ground truth, predicted binary images with threshold 0.75, 
        refined binary images, binary images generated from the 
        ordinary semantic segmentation method (``traversable plant'' pixels
        are shown in white), and binary images generated from 
        the probability maps predicted by the baseline method \cite{Kim2007}.
        Note that the traversable regions are not 
        identical to the plant regions, but plant parts such as 
        stems are not traversable.
        TEM is capable of distinguishing the traversable plants from the non-traversable ones
        (highlighted in yellow ellipses).
        The refinement by predicted object classes mainly worked as noise suppression
        (highlighted in red rectangles).
        }
    \label{fig:tem_visualization}
\end{figure*}

We evaluate the binary traversability images 
generated from the predicted traversability
with different thresholds.
Here we compare two types of binary images: {\it raw} and {\it refined}.
The {\it raw} binary images are 
generated by just binarizing the predicted traversability with a threshold.
The {\it refined} binary images are
generated by additionally setting all the pixels predicted as 
an artificial object or the ground to non-traversable.
This refinement provides the filtering effect of false positive 
traversability predictions even in the case of high predicted traversability,
which is equivalent to the calculation in 
the voxel described in \ref{sec:proposed_method_3d_semantic_voxel_map}
where the final decision of traversability is given considering 
both the object class and the predicted traversability.
Fig. \ref{fig:pr_curve} shows a Precision-Recall curve
as well as an IoU-Recall curve to analyze the performance of TEM.
The best IoU was achieved when the threshold was 0.75.
Therefore, we use this traversability threshold in the rest of the paper.
Table \ref{table:performance_of_TEM} shows the accuracy, precision,
and recall of the model on the test set with
the threshold of 0.75.

From Fig. \ref{fig:pr_curve} and Table \ref{table:performance_of_TEM},
we can see high recall and relatively low precision.
The refined predictions resulted in better precision and IoU.
This result indicates that fusing
the predictions of object classes and traversability
decreased the false positive rate and thus is suitable for safe robot operation
compared to only predicting the traversability.

Fig. \ref{fig:tem_visualization} is the visualization of
the prediction by TEM.
Much of the traversable regions are classified as positive.
In addition, some non-traversable plants are predicted correctly.
This result shows that our model can learn the differences between
traversable and non-traversable plants.
In terms of the refined results, we cannot qualitatively see obvious
improvements, but we confirmed that small prediction noise was suppressed.
Note that these segmentation results are yielded from automatically generated
traversability masks, which are incomplete and noisy.
Even from such data, our PU learning-based training framework effectively
learns the features of traversable objects, and provides reasonable estimation.

\noindent
\textbf{Comparison to the semantic segmentation}
TEM provided better performance than 
the segmentation-only method in all the metrics.
Especially, the IoU and the recall were significantly low
in the ordinary semantic segmentation.
In Fig. \ref{fig:tem_visualization}, we can see that
the semantic segmentation method could not capture
the traversable plant regions well.
We suppose it is because the labeling of ``traversable plant'' regions
does not cover all the traversable plant regions as mentioned
in \ref{sec:traversability_mask} and results in inconsistent labels
with traversable plant regions labeled as non-traversable plant.
Such labels confuse the training.
Therefore, the segmentation-only method
does not get along with the sparse traversable masks.
On the other hand, our proposed method enables more accurate and 
effective training of the model to 
estimate traversability of image regions.

\noindent
\textbf{Comparison to Kim et al.}
We report the average of the metrics over the 10 different
partitions for the cross-validation
in Table \ref{table:performance_of_TEM}.
For each data partition, the threshold of the probability 
is determined so that it yields the best mean IoU for the test images.
As a result, our proposed method outperformed the baseline in all the metrics
even though the baseline model is trained with ground truth labels 
while our method is trained with noisy and incomplete traversability masks.
As can be seen in Fig. \ref{fig:tem_visualization},
although the plant regions are mostly classified correctly,
other regions of artificial objects and the ground are also classified 
as traversable.
Moreover, the traversable leaves and branches
and non-traversable stem parts are not distinguished and 
almost equally classified as traversable.

Our method is advantageous over the baseline by Kim et al. in three aspects.
First, our PU learning-based method 
enables the training without negative labels that are
difficult to acquire.
In \cite{Kim2006}, negative labels are collected using the information 
of the robot actually bumping into an obstacle, which is
not appropriate in environments such as greenhouses.
Our learning framework allows for safer and more practical way of training
the scene recognition model by utilizing only positive labels 
acquired via the control by a human operator.
Second, our model provides semantics about object classes
to complement the estimation of traversability.
As can be seen in Fig. \ref{fig:tem_visualization},
there are a lot of regions wrongly classified as traversable
in the prediction of the baseline method.
There is no way to correct the misclassifications in the baseline,
while our method can amend them using the prediction by SSM as mentioned above.
Third, our method can utilize more discriminative features
thanks to the DNN model trained in the task of semantic segmentation,
which led to the better performance of the proposed method over the baseline
even by the raw predictions of TEM.

\subsection{Navigation in a greenhouse}
\label{sec:navigation_in_a_greenhouse}

\begin{figure}[tb]

  \subfloat[Robot configuration\label{fig:robot_configuration}]{
  \includegraphics[width=0.45\hsize]{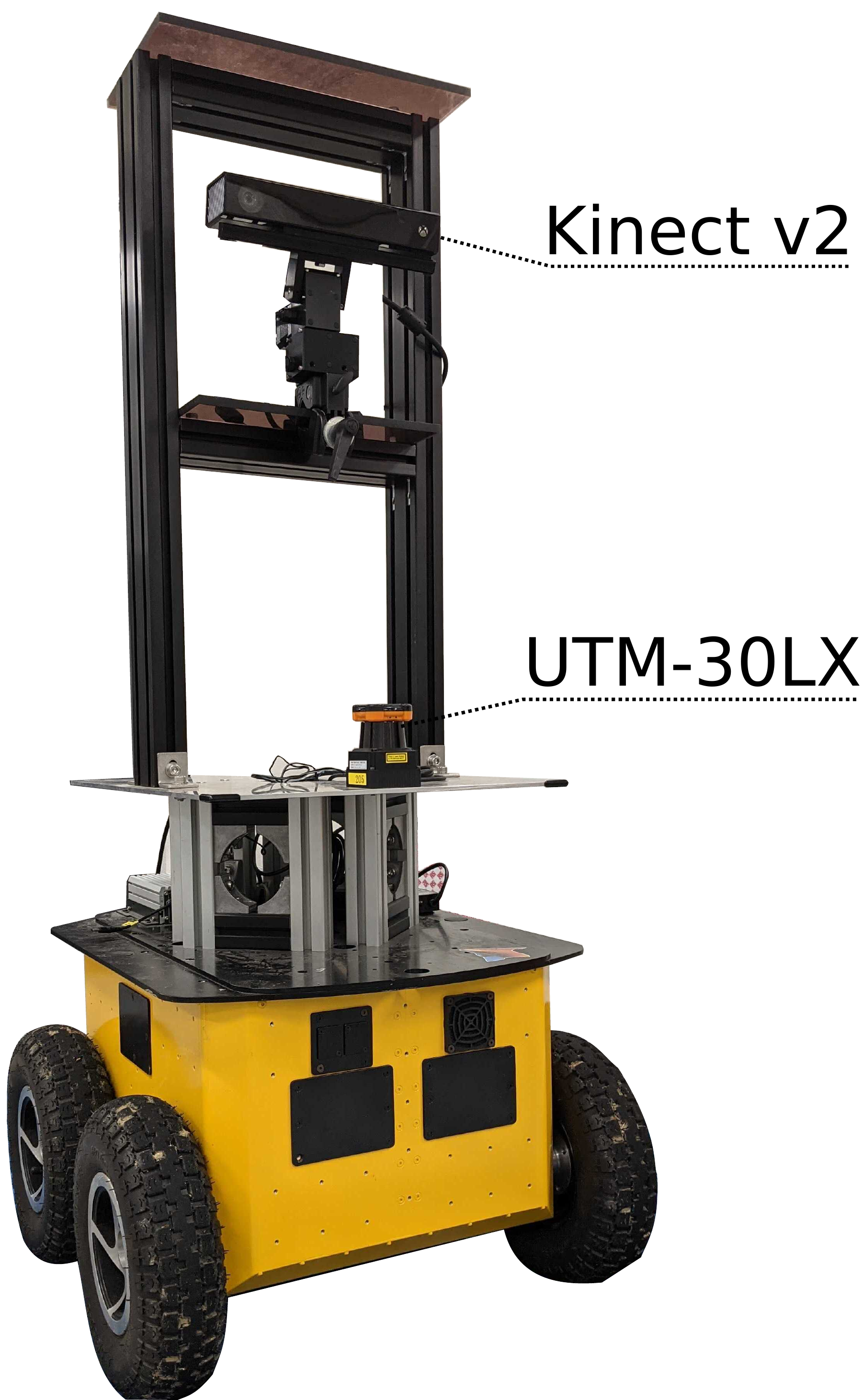}}
  \subfloat[Inside the greenhouse\label{fig:inside_greenhouse}]{
  \includegraphics[width=0.45\hsize]{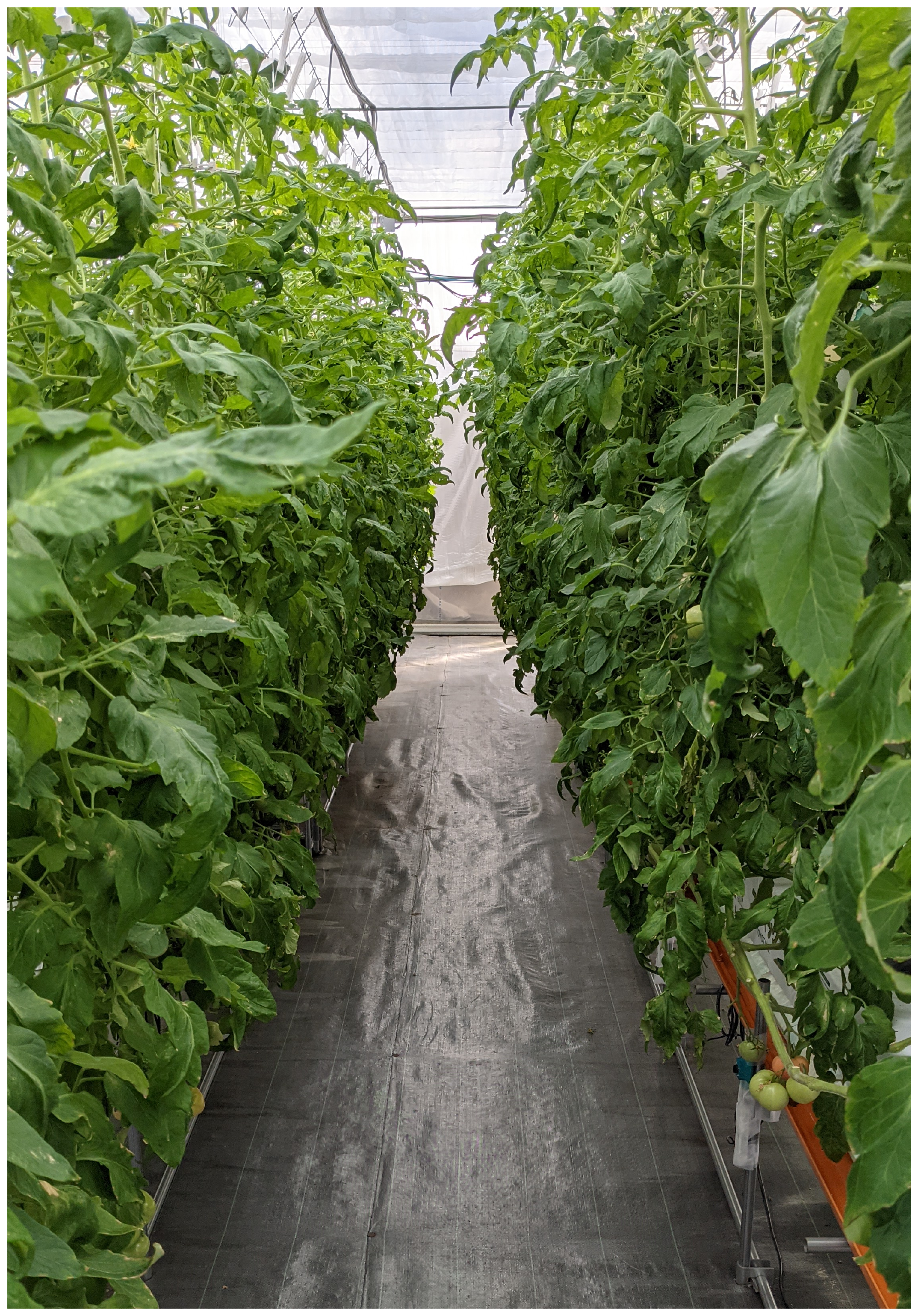}}
    \caption{Experiment environment}
    \label{fig:experiment_environment}
\end{figure}

Finally, we conducted a real-world navigation experiment in a greenhouse.
All the software is processed on
a laptop with an Intel Core i7-6700HQ, an NVIDIA GeForce 960M, and 32 GB RAM.
Revast Mercury is used as a robot platform (See Fig. \ref{fig:robot_configuration}).
As sensors, the robot is equipped with 
a Kinect v2, an RGB-D sensor, and Hokuyo UTM-30LX, a laser range finder.

An image of a path in the greenhouse is shown in Fig. \ref{fig:inside_greenhouse}.
The length of the paths is approximately 7.5 [m].
The software for the DNN prediction is implemented on
Robot Operating System (ROS) \cite{Quigley2009}
with C++ and Libtorch, a C++ library for PyTorch.
The frame rate of the prediction of the network is approximately 8 [fps]
and the 3D mapping runs at approximately 5 [fps].
The frame rate of the original ESPNetv2 
is also approximately 8 [fps]  on the same computer and thus 
our modification to the network does not affect 
the computational efficiency.

As a baseline for navigation,
we also test a system that considers all the voxels as obstacles.
It corresponds to using only geometric information,
which is a major approach in most of the current mobile robots. 

\begin{figure*}[tb]
  \subfloat[\label{fig:map_1}]{
  \includegraphics[width=0.33\hsize]{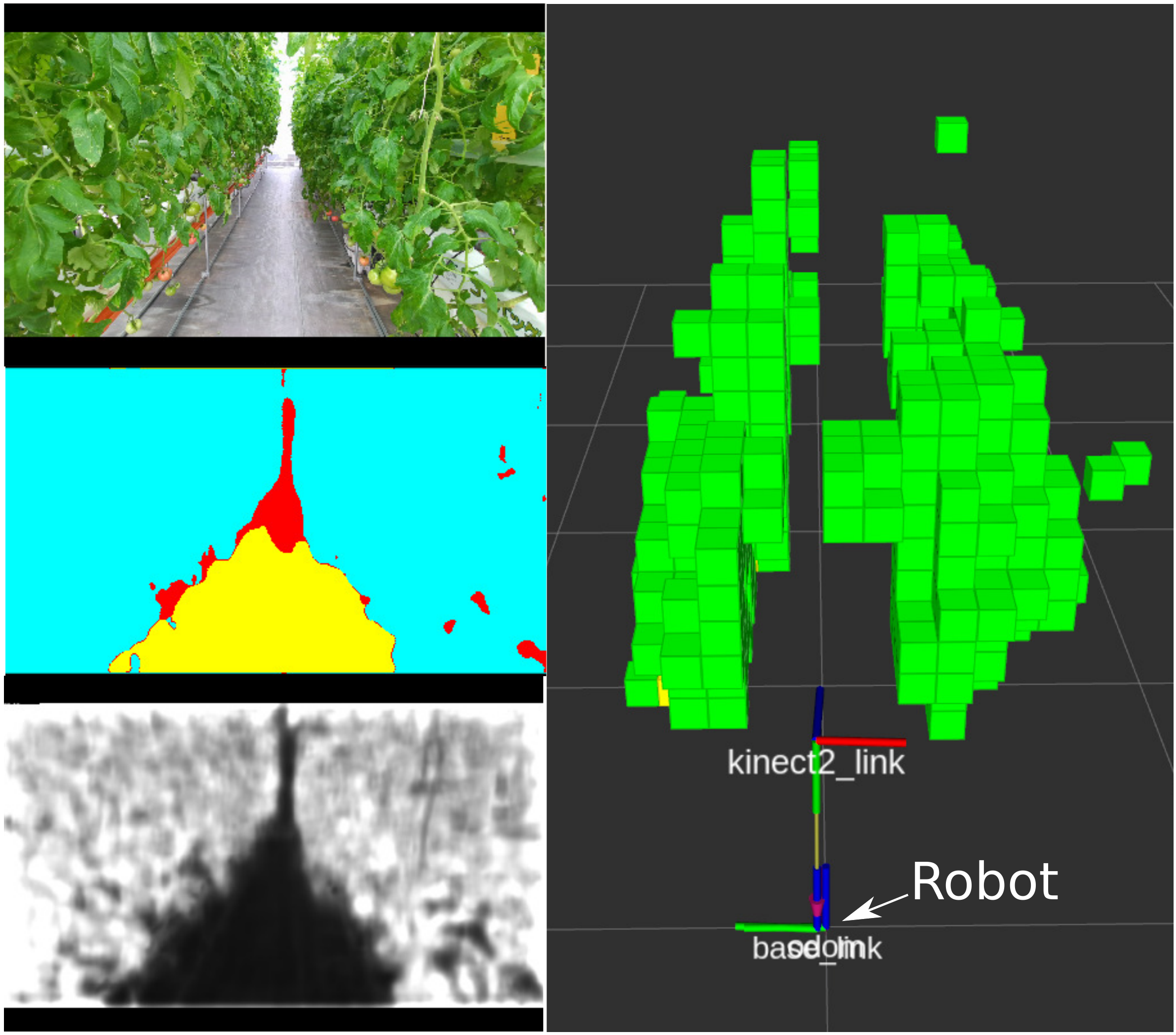}}
  \subfloat[\label{fig:map_2}]{
  \includegraphics[width=0.33\hsize]{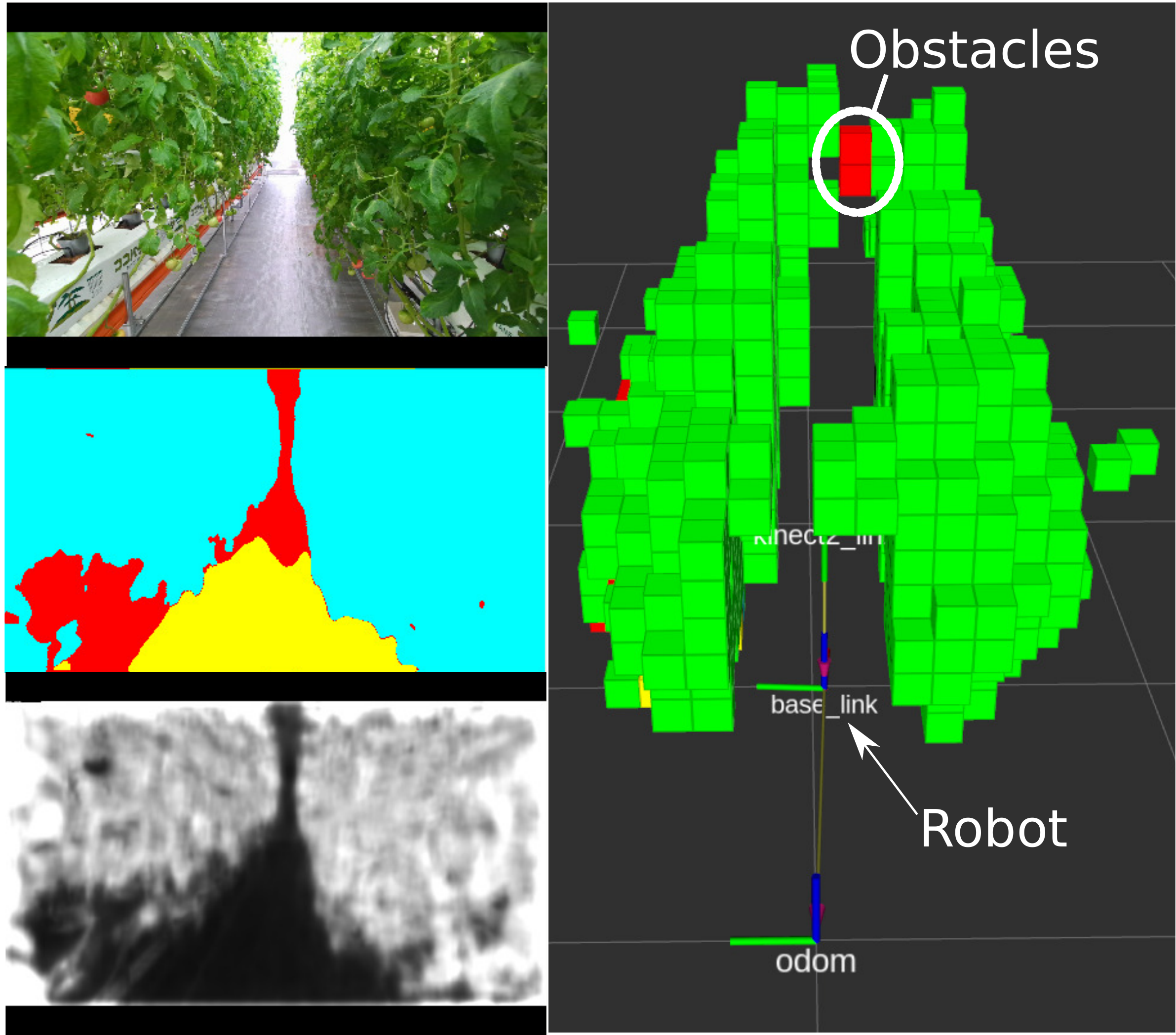}}
  \subfloat[\label{fig:map_3}]{
  \includegraphics[width=0.33\hsize]{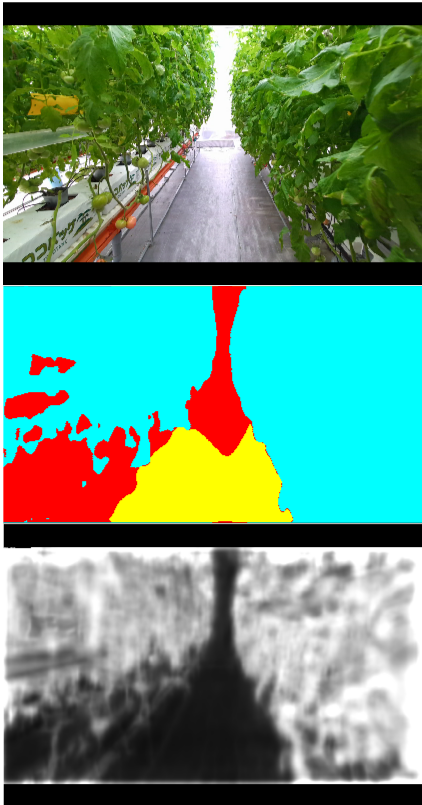}}
  
    \caption{Generated 3D maps during the experiment.
    An input image, a predicted segmentation image,
    and a predicted pixel-wise traversability are shown on the left side of each figure.
    In the segmentation images, {\it blue} is the plant, {\it red} is the artificial object
    and {\it yellow} is the ground.
    In the figures of 3D map, {\it green} indicates traversable voxels with the plant class
    and the traversability higher than the threshold.
    The rest of the colors indicates the same object classes as in the segmentation images.}
    \label{fig:experiment_map}
\end{figure*}

\subsubsection{Obstacle detection in a simple forward motion}
\label{sec:navigation_constant_control}

As a primary experiment, we conducted an experiment of 
obstacle detection using the proposed 3D semantic map
during the robot's act of moving forward.
We assume that the path is straight and
constant control signals of linear velocity are given to the robot while
the robot recognizes the traversability of objects in front.
The linear velocity is set to 0.1 [m/s].
An obstacle point cloud is generated from the 3D semantic voxel map
by outputting only voxels other than ``traversable plant'' as obstacle points.
When an obstacle point is detected within a range in front of the robot,
the control signal is stopped.

%
%
We carried out 5 trials on the same path. 
The baseline navigation method consistently failed the task
due to the part of plants grown out to the path.
In contrast, our system was able to recognize the traversable plants and
navigate through the path between the plant rows in all the trials.
This result shows the ability of our scene recognition method 
to recognize the traversable plants covering the path.
There were, however, some cases where the robot stops for a while and then restarts
even though there was no obstacle.
This was due to misclassification of voxels in front of the robots as obstacles.
The misclassification of voxels stems either from the error in the image-based
semantic segmentation or from the registration error between the predicted segmentation images
and the corresponding depth images, which RGB-D sensors inevitably have.
Errors temporarily occurring on a few pixels are suppressed by the voxelization 
and the Bayes' update.
When the error consistently occurs on a large area, however,
the corresponding voxel can be misclassified.

Fig. \ref{fig:experiment_map} is the visualization of the semantic 3D maps
generated during the experiment at different times in chronological order.
There were a lot of plant parts partially covering the path.
Our proposed method was able to recognize the traversable plants 
and to traverse them.
Fig. \ref{fig:map_2} shows the case where the system wrongly recognized 
the regions in a far front of the robot as obstacles shown in the red voxels. 
The robot stopped moving in response to the obstacles.
After observing a few data frames while stopping,
the obstacle voxels were removed from the map
and the robot resumed the navigation.
We suppose the cause of this phenomenon as follows.
When the misclassified regions are far from the robot,
the segmentation results and the depth readings can be more noisy
and it results in the misclassification of the voxels
or creating voxels on wrong locations.
In the case in Fig. \ref{fig:map_2}, it is more likely that 
the voxels were spawned on wrong locations due to the depth noise.
As the robot approaches the voxels, both the segmentation and the depth values 
become more accurate and the voxels are updated with those observations.
As a result, in the case in Fig. \ref{fig:map_3},
the misclassified voxels were deleted from the voxel map
because points in those voxels were not observed for the pre-defined
number of frames (10 frames)
when the robot was stopping near them.


\subsubsection{Integration with move\_base}
\label{sec:navigation_move_base}

We also conducted an experiment of applying our 3D semantic voxel map to navigation using
move\_base \cite{move_base_web}, a de facto standard navigation software in ROS \cite{Quigley2009}.

The obstacle point cloud is fed in move\_base as a sensor reading.
In addition to the obstacle point cloud, a 2D laser range finder (LRF) is used 
to complement the blind spot of the RGB-D sensor and stabilize the navigation.
Because of this purpose, the obstacle range of the LRF is limited to 1.0 [m].
The configuration of the LRF is the same in the baseline method.
Sub-goals are given manually via Rviz,
an interactive GUI tool for visualization.


Here, we define criteria for judging the success or failure of the trials.
In all the trials, we observed that the accuracy of the SSM 
consistently degrades about 2.5 [m] before the end of the paths
and the robot fails to detect the traversable path
(We discuss this problem in \ref{sec:discussion}).
We therefore consider a trial in which the robot reached about 2.5 [m] 
before the end of the path as ``traversed``.
When the robot stops due to misclassification,
we allow a human's intervention by
manually restarting move\_base to clear the costmap at most once.
Trials where the robot gets stuck even after the human intervention are
marked as ``stuck''.

\begin{figure*}[tb]
  \subfloat[Input image\label{fig:move_base_input}]{
  \includegraphics[width=0.390\hsize]{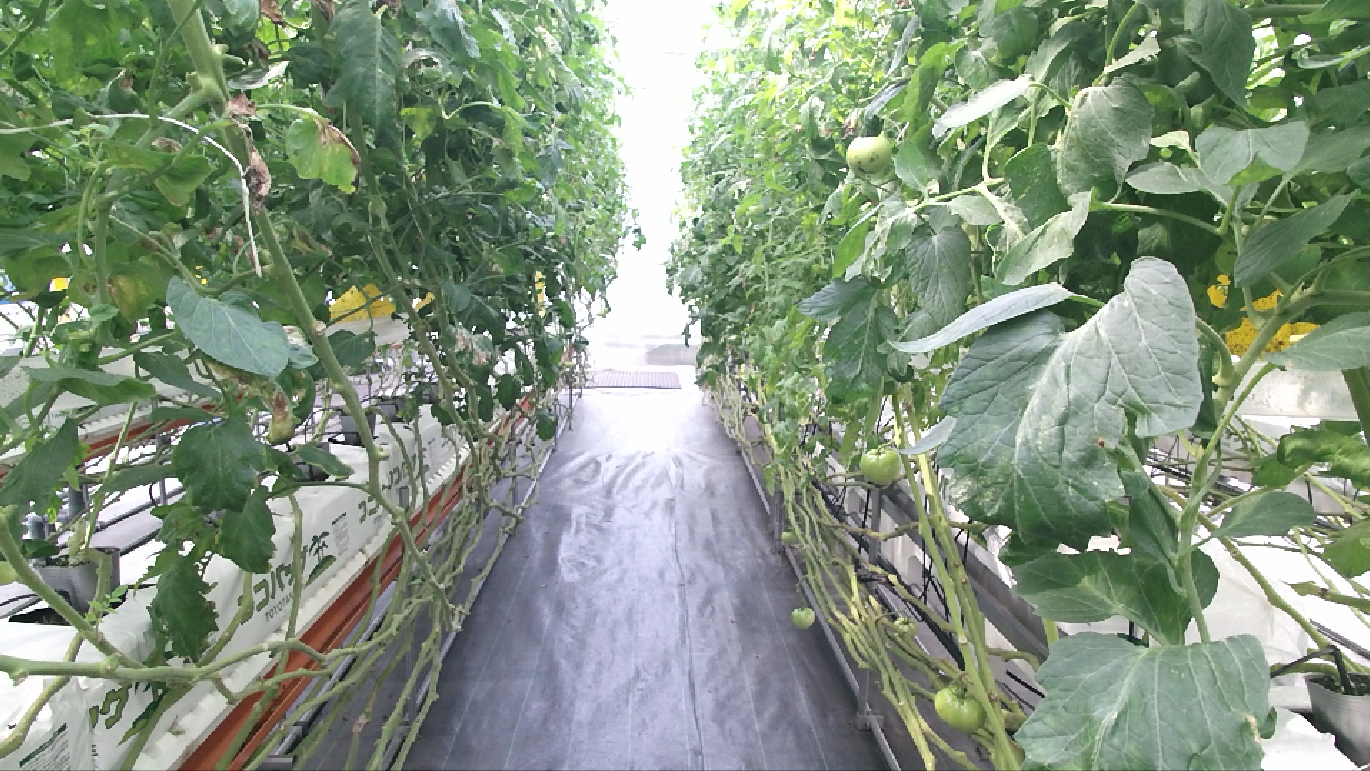}}
  \subfloat[Baseline for navigation\label{fig:move_base_baseline}]{
  \includegraphics[width=0.305\hsize]{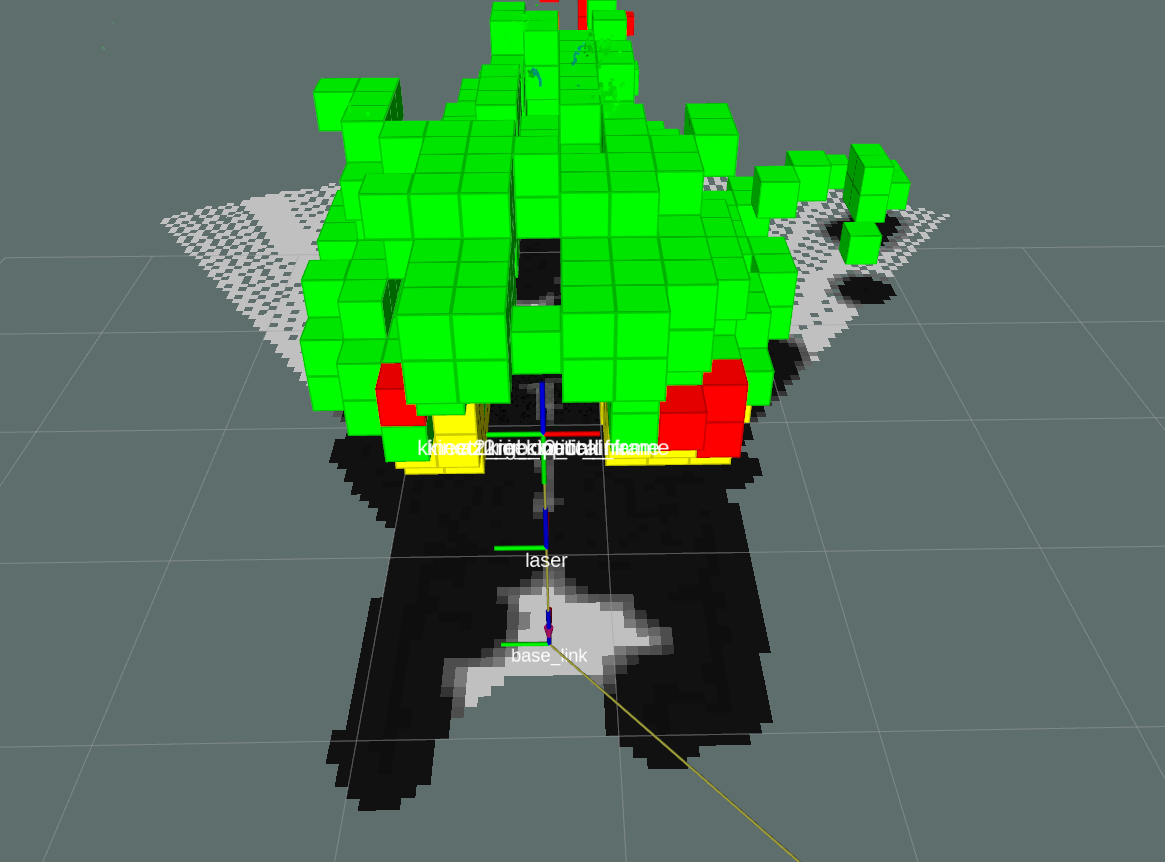}}
  \subfloat[Ours\label{fig:move_base_trav_map}]{
  \includegraphics[width=0.305\hsize]{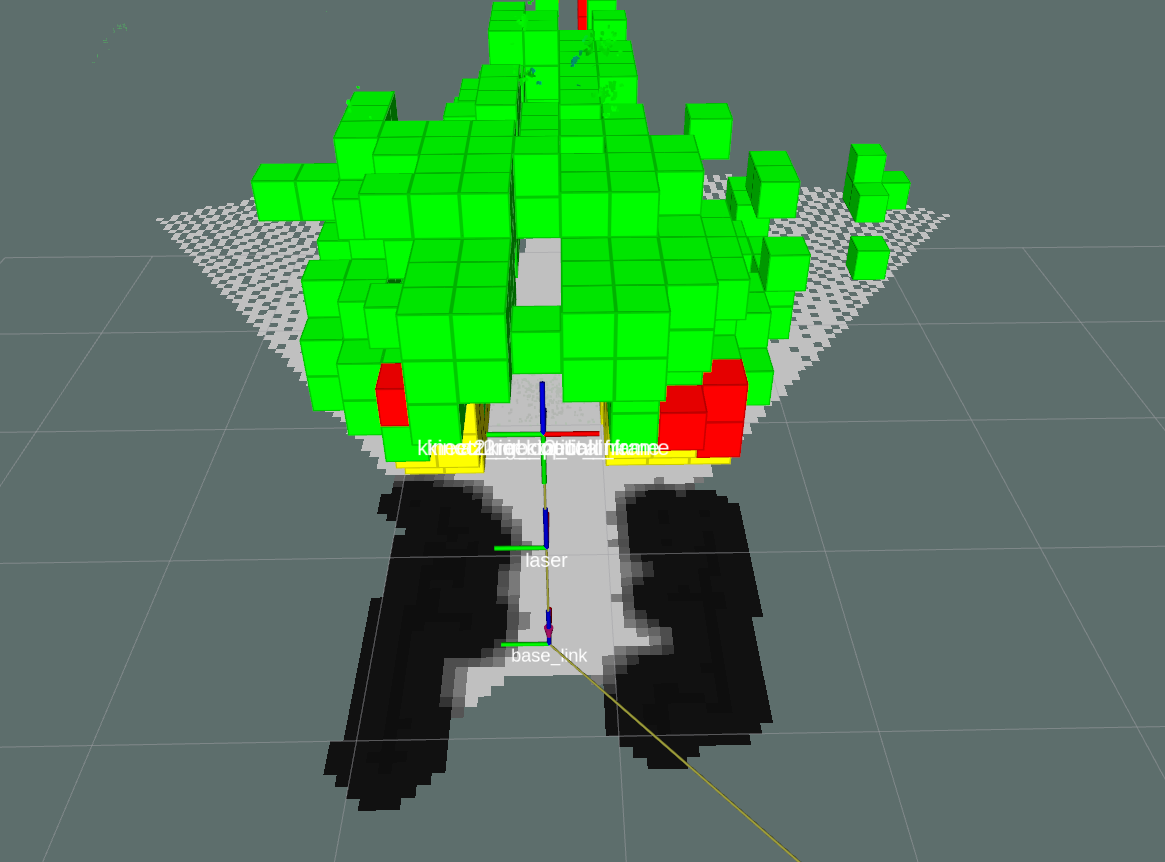}}
  
    \caption{Costmaps in the baseline and the proposed method. 
    (a) An input image. (b) Semantic 3D voxel map and the costmap of the baseline.
    All the voxels are considered as obstacles and thus the path is recognized 
    as blocked.
    (c) Semantic 3D voxel map and the costmap generated by the proposed method.
    The voxels of traversable plant are not considered as obstacles and 
    the path is recognized as traversable.}
    \label{fig:move_base_cost_map}
\end{figure*}

The baseline navigation method consistently failed 
to find a path on three paths
before entering them.
An example of a costmap generated with the baseline method
is shown in Fig. \ref{fig:move_base_baseline}.
We carried out 12 trials of navigation on the three paths (4 each)
with the proposed method.
The result is shown in Table \ref{table:result_navigation_move_base}.
\begin{table}
    \centering
    \caption{Result of the navigation experiment using {\it move\_base}}
    \label{table:result_navigation_move_base}
    \begin{tabular}{ccc} \toprule
     Trials & Traversed (intervened) & Stuck \\ \midrule
     12 & 8 (5)     & 4         \\ \bottomrule
    \end{tabular}
\end{table}
The robot with the proposed method succeeded 
to traverse through the path partially covered by plants in 8 trials,
5 of which involved a human's intervention.
The costmaps generated with the proposed method 
are compared in Fig. \ref{fig:move_base_trav_map}.
From these results, we can see that our proposed method enables 
the robot to recognize the traversable plant parts, which would otherwise 
be recognized as obstacles, and traverse through them.
However, misclassifications affected the robot's motion 
as can be seen from the number of required interventions and failures.

\subsection{Discussion}
\label{sec:discussion}

\subsubsection{Cause of the failures of navigation}
We showed that our method enabled navigation through the paths 
covered by traversable plant parts in the greenhouse,
while the baseline method constantly failed.

In all the trials, however, the robot with our proposed method 
consistently failed to navigate 
about 2.5 [m] before the end of the paths.
In addition, in some of the successful trials, 
it required a human's intervention due to misclassifications.
On the other hand, in the experiment in \ref{sec:navigation_constant_control},
all the 5 trials were successful without any interventions.

We suppose that the difference of the degrees of
the success rates stems from the difference of the density of leaves.
In such a case, majority of the regions are dominated by a single class of ``plant''
and a few misclassifications are filtered out in the probability calculation in 
each voxel.
In the experiment in \ref{sec:navigation_move_base}, however,  
leaves were sparse and artificial objects behind the plant rows were also visible
through the leaves. 
This was especially obvious near the end of the path.
Since the SSM is trained with pseudo-labels, which are generated by merging 
the outputs from multiple pre-trained models, the model lacks pixel-level accuracy.
This is the current limitation of our method.

For more reliable navigation, the accuracy of the recognition model 
needs to be improved.
In addition, to deal with the prediction noise in DNN models, 
a better filtering and/or refinement methods should be implemented.

\subsubsection{Adaptability of the proposed system}

The success of the training of our scene recognition model 
depends on the UDA of semantic segmentation, 
since TEM takes the intermediate features of the semantic segmentation network
(SSM) for estimation.
In terms of the segmentation tasks in greenhouses,
we confirmed in the previous work \cite{Matsuzaki2020}
that our UDA method is applicable to multiple greenhouses 
and multiple seasons.
The proposed method will thus also be applicable to various 
greenhouse environments.
When applying the proposed system in a new environment, in general,
the segmentation model will need to be newly trained to adapt to 
the environment.
Once the segmentation model is trained,
TEM can be trained in the same manner as described in this paper.
Note that our proposed model and the training method
provide advantages of the training with a little images and incomplete 
positive labels, which lead to a minimum burden in deployment.

\subsubsection{Influence of the sensor setting}

In this work, we employed a Kinect v2 for simplicity of implementation
of the sensor system to acquire calibrated RGB and depth images.
Although it is sensitive to disturbance from strong light,
we did not observe any severe effect of the sunlight on the depth readings.
in our experiments.
We suppose it is because the sunlight was cut by the translucent roof and walls 
of the greenhouse and the leaves, and thus 
the sensor was able to work in moderate lighting conditions.
When we apply the system to outdoor environments, however,
the choice of sensors will influence the accuracy and reliability of the system.
Note that our proposed method is compatible with any sensors
that can provide registered RGB images and depth readings
and the sensors can be selected based on the lighting condition of 
the target environment.
Applicable sensor settings include
a stereo sensor, and well-calibrated 3D LiDAR and a monocular camera.

In terms of sensor calibration, Kinect v2 was used out-of-the-box in our experiments.
In multi-sensor setting, it will indeed require a precise calibration,
although the Bayes' update and the voxelization process can to some extent deal with 
a little noise due to the misalignment of the sensors.

\section{CONCLUSIONS}

In this paper, we described a method of estimating the traversability of
plants covering a path and navigating through them
in plant-rich environments for mobile robots.
We proposed the following novel methods.
1. A scene recognition model that estimates
general object classes and the traversability of the objects.
2. A manual annotation-free training of the model
using an unsupervised domain adaptation method
for the semantic segmentation module (SSM) 
and the information of the robot's traversals,
named \textit{Traverasability masks}
for the traversability estimation module (TEM).
3. A PU learning-based training method
to effectively train TEM with the traversability masks, which include 
unlabeled traversable regions.
Our proposed method enables easy and safe deployment 
of mobile robots with the capability of recognizing traversable plants 
with a minimum burden.
In the comparative analysis, we confirmed that the proposed method 
is able to estimate the traversability of image regions more accurately 
than a conventional semantic segmentation and an existing work of 
image-based traversability estimation.
Moreover, we applied the scene recognition model to 
the navigation tasks in a real greenhouse with plants covering the paths
and confirmed its ability to recognize traversable plants 
and navigate through them.

As future work, we are going to work on the improvement of the navigation system.
In particular, the accuracy of the scene recognition and the filtering of 
misclassification should be improved.
We are also looking to applying this work to a navigation task
in unstructured environments such as forest paths.

\appendices

\section{Details of the network structure}

\begin{figure*}[tb]
  \centering
  \subfloat[Network architecture\label{fig:network_detail_entire_network}]{
  \includegraphics[width=0.45\hsize]{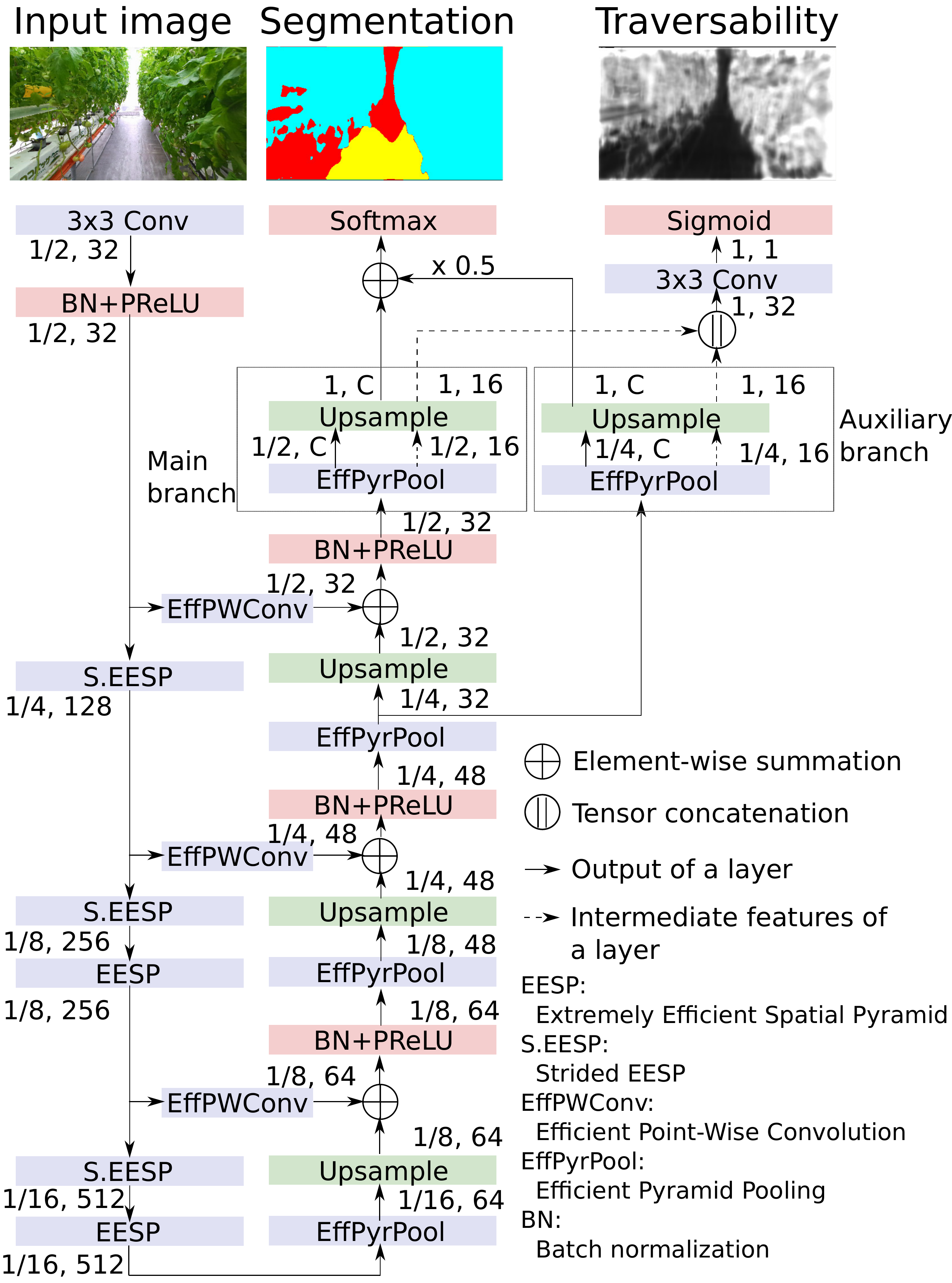}}
  \subfloat[Efficient Pyramid Pooling (EffPyrPool)\label{fig:network_detail_effpyrpool}]{
  \includegraphics[width=0.52\hsize]{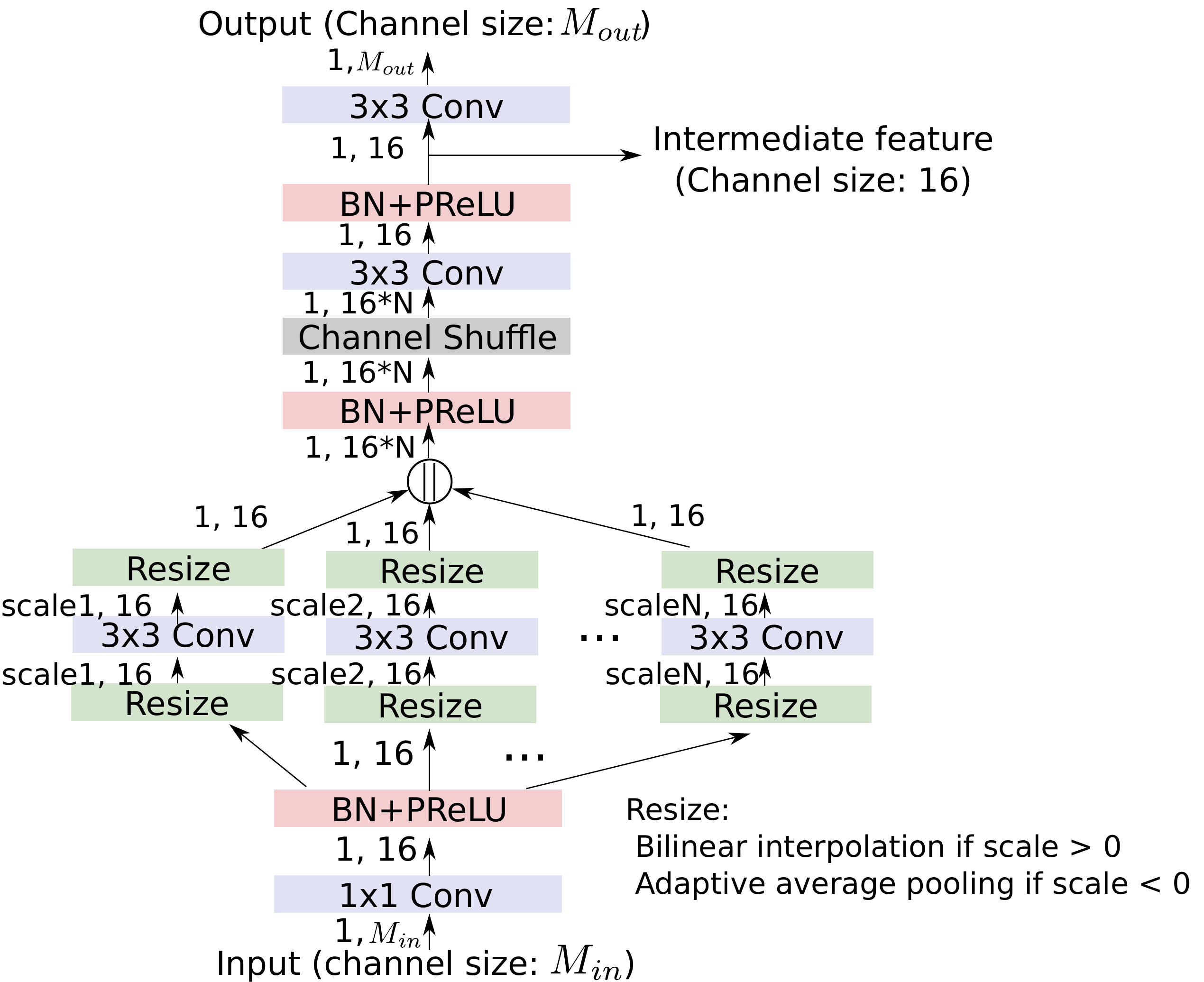}}
  
  \caption{A detailed diagram of our architecture.
  The numbers beside each layer 
  shows the spatial resolution relative to the input image and 
  the channel size of the output of the layer, respectively.
  Based on ESPNetv2 \cite{Mehta2019}, an auxiliary segmentation branch 
  is added to estimate pixel-wise uncertainty for the training of SSM
  proposed in \cite{Matsuzaki2020}. For estimating traversability,
  the intermediate features of the two segmentation branches
  are concatenated and fed into a $3\times3$ convolution followed by 
  the sigmoid function. For the detail of each network layer,
  refer to \cite{Mehta2019}.}

  \label{fig:network_detail}
\end{figure*}

Fig. \ref{fig:network_detail_entire_network} shows a complete
structure of our network architecture.
Our network is based on ESPNetv2 \cite{Mehta2019},
a light-weight semantic segmentation network.
It consists of modules such as Extremely Efficient Spatial Pyramid (EESP),
Efficient Point-Wise Convolution (EffPWConv),
Efficient Pyramid Pooling (EffPyrPool),
which consist of several convolutions with different types of kernels,
batch normalization, and activation layers.
For detailed descriptions of those modules, 
we refer the readers to \cite{Mehta2019}.
An auxiliary segmentation branch
and a branch for estimating traversability, i.e., TEM
are attached 
to the middle of the main network of ESPNetv2. 
The features for TEM are generated by concatenating 
the intermediate features from the EffPyrPool layer in each segmentation branch
(See Fig. \ref{fig:network_detail_effpyrpool}).
The intermediate feature from the SSM is 
fed into a $3\times3$ convolution and
the sigmoid function to produce pixel-wise probability predictions.

\printbibliography

\begin{IEEEbiography}[{\includegraphics[width=1in,height=1.25in,clip,keepaspectratio]{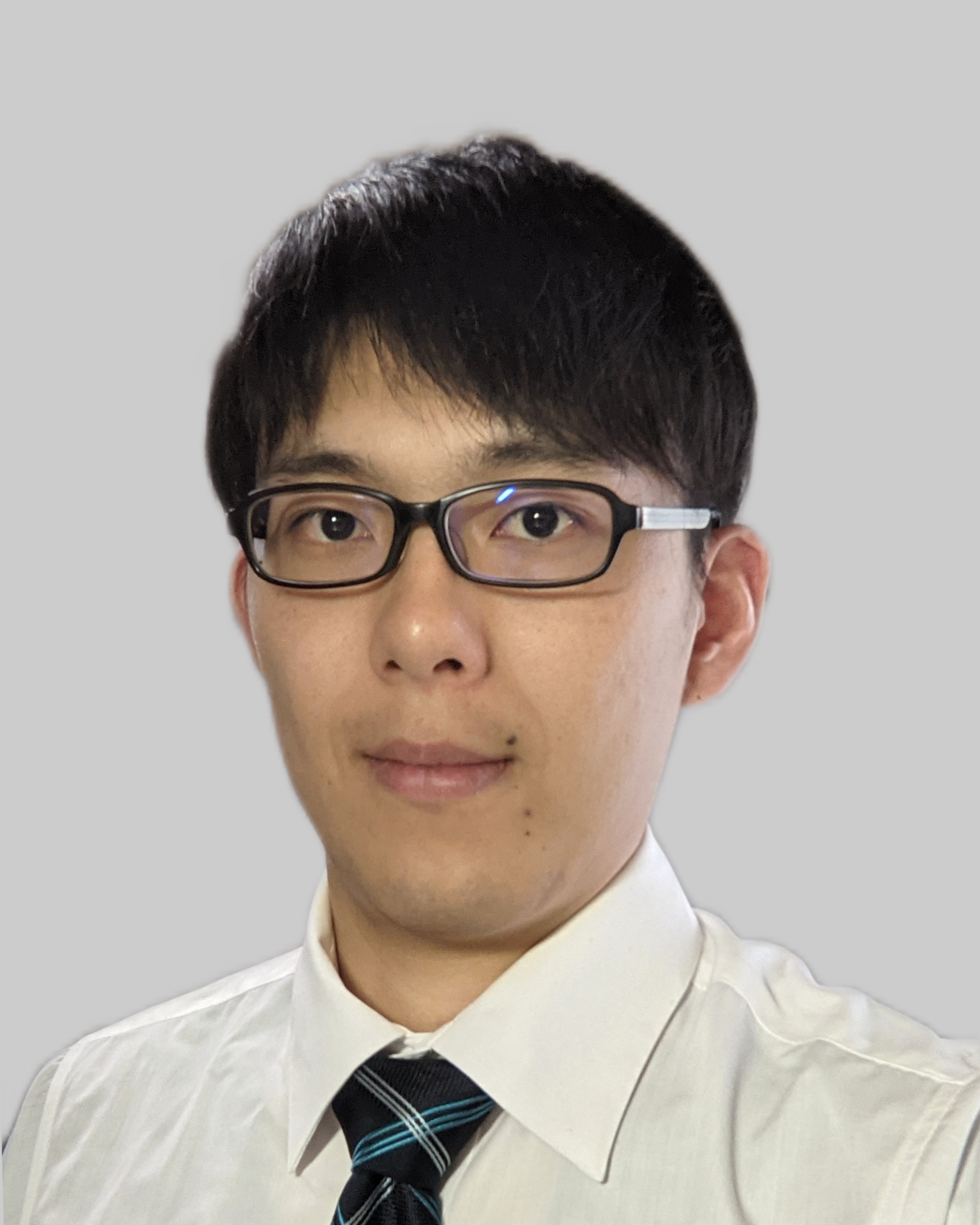}}]{Shigemichi Matsuzaki}
  (Graduate Student Member, IEEE)
  received the associate degree in engineering from
  National College of Technology, Kumamoto College in 2016,
  and the bachelor's and the master's degrees in engineering
  from Toyohashi University of Technology, Aichi, Japan
  in 2018 and 2020, respectively.
  He is currently pursuing the Ph.D. degree 
  at Toyohashi University of Technology.
  His research interests include deep learning in robot perception,
  mobile robots in unstructured environments, and mobile service robots.
\end{IEEEbiography}

\begin{IEEEbiography}[{\includegraphics[width=1in,height=1.25in,clip,keepaspectratio]{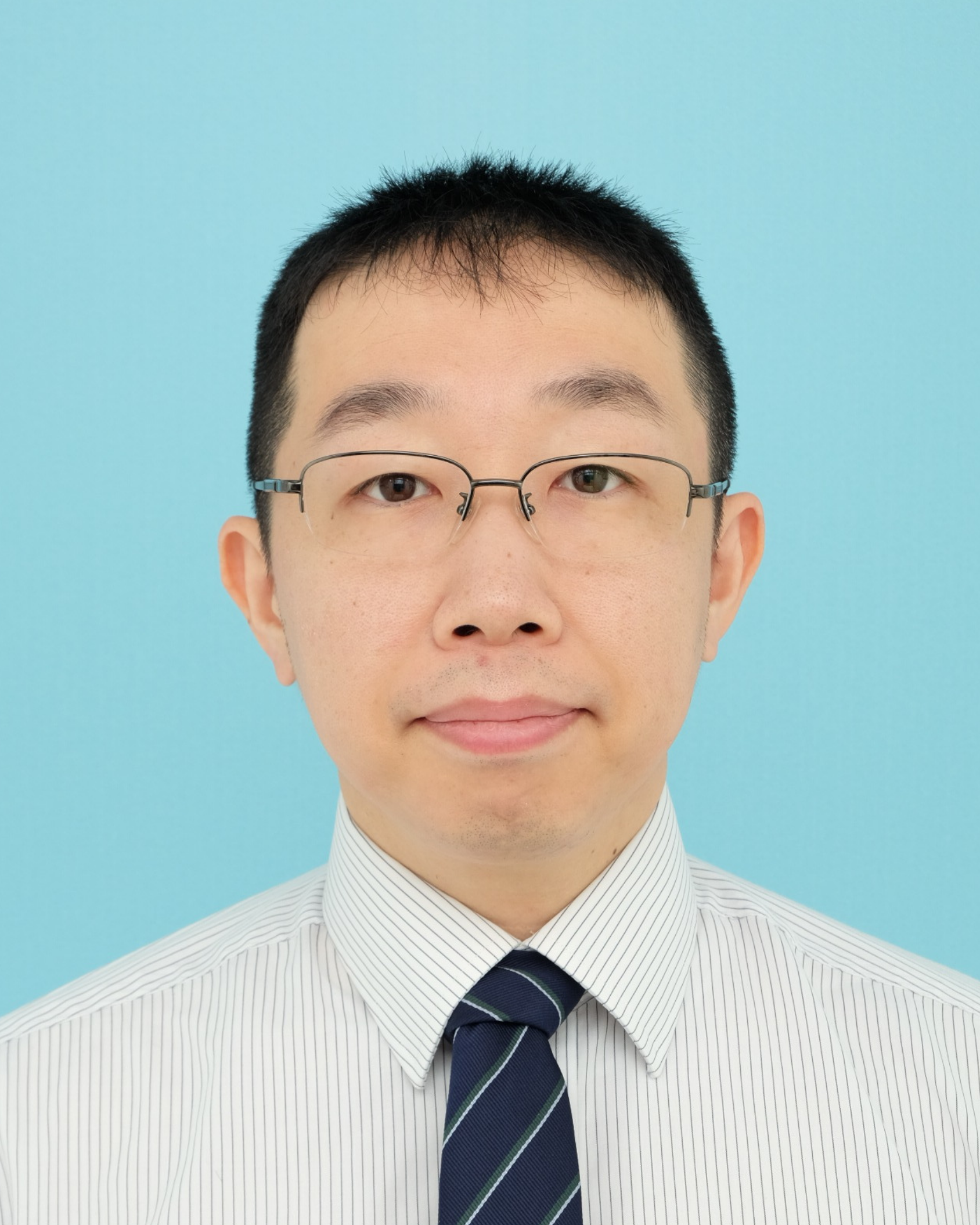}}]{Hiroaki Masuzawa}
  received the associate degree in engineering from
  National College of Technology, Hachinohe College in 2006,
  received the bachelor's and the master's degree in engineering from
  Toyohashi University, Aichi, Japan in 2008 and 2010, respectively.
  He is currently a project
  research associate of Department of Computer Science and Engineering,
  Toyohashi University of Technology.  His research interests include
  application of artificial intelligence to agricultural robots.
\end{IEEEbiography}

\begin{IEEEbiography}[{\includegraphics[width=1in,height=1.25in,clip,keepaspectratio]{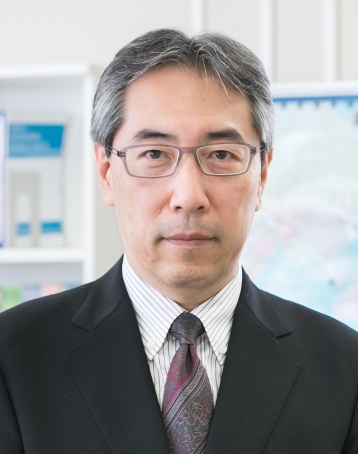}}]{Jun Miura}
  (Member, IEEE)
  received the B.Eng. degree in mechanical engineering, and
  the M.Eng. and Dr.Eng. degrees in information engineering from The University of Tokyo,
  Tokyo, Japan, in 1984, 1986, and 1989, respectively. In 1989, he joined the Department of
  Computer-Controlled Mechanical Systems, Osaka
  University, Suita, Japan. Since April 2007, he has
  been a Professor with the Department of Computer
  Science and Engineering, Toyohashi University of
  Technology, Toyohashi, Japan. From March 1994 to February 1995, he was a
  Visiting Scientist with the Computer Science Department, Carnegie Mellon
  University, Pittsburgh, PA. He has published over 220 articles in international
  journal and conferences in the areas of intelligent robotics, mobile service
  robots, robot vision, and artificial intelligence. He received several awards,
  including the Best Paper Award from the Robotics Society of Japan, in 1997,
  the Best Paper Award Finalist at ICRA-1995, and the Best Service Robotics
  Paper Award Finalist at ICRA-2013
\end{IEEEbiography}

\EOD

\end{document}